\pdfoutput=1
\documentclass[11pt]{article}

\usepackage[preprint]{acl}

\usepackage{times}
\usepackage{latexsym}
\usepackage{booktabs}
\usepackage{multirow}
\usepackage{tabularx}
\usepackage{amsmath}
\usepackage{xcolor}
\usepackage{color, colortbl}
\definecolor{Gray}{gray}{0.9}
\usepackage{mdframed}
\usepackage{algorithm}
\usepackage{algpseudocode}
\usepackage[english]{babel}
\usepackage{amsthm}

\theoremstyle{definition}
\newtheorem{definition}{Definition}

\theoremstyle{remark}

\usepackage[T1]{fontenc}
\usepackage[utf8]{inputenc}

\usepackage{microtype}

\usepackage{inconsolata}

\usepackage{graphicx}

\usepackage{listings}
\usepackage{xcolor} % For color support

\usepackage{times}
\usepackage{microtype}
\usepackage{adjustbox}

\usepackage{enumitem}
\usepackage[most]{tcolorbox}
\tcbset{aibox/.style={
    breakable,
    break at=\maxdimen,
    fontlower=\scriptsize, 
}}
\newtcolorbox{AIbox}[2][]{aibox,title={#2},#1}

\newcommand{\jx}[1]{\color{black}{#1}}

\title{AutoSchemaKG: Autonomous Knowledge Graph Construction through Dynamic Schema Induction from Web-Scale Corpora}

\author{Jiaxin Bai\thanks{~~Equal contribution.}$^1$, Wei Fan\footnotemark[1]$^1$, Qi Hu\footnotemark[1]$^1$, Qing Zong$^1$, Chunyang Li$^1$, Hong Ting Tsang$^1$ \\\textbf{Hongyu Luo$^1$, Yauwai Yim$^1$, Haoyu Huang$^2$, Xiao Zhou$^1$, Feng Qin$^1$, Tianshi Zheng$^1$} \\\textbf{Xi Peng$^3$, Xin Yao$^3$, Huiwen Yang$^3$, Leijie Wu$^3$, Yi Ji$^3$} \\\textbf{Gong Zhang$^3$, Renhai Chen$^3$, and Yangqiu Song$^1$}\\
  $^1$CSE, HKUST 
  $^2$ CSE, CUHK
  $^3$ Theory Lab, Huawei \\
  \texttt{\{jbai, wfanag, qhuaf, qzong, cliei, httsangaj, hluoay, ywyimaa, tzhengad} \\\texttt{xzhoucs, fqinac\}@cse.ust.hk}, \texttt{ haoyuhuang@link.cuhk.edu.hk } \\
   \texttt{ \{pancy.pengxi, yao.xin1, huiwen.yang, leijie.wu1, jiyi13,}\\\texttt{ nicholas.zhang, chenrenhai\}@huawei.com }  \texttt{yqsong@cse.ust.hk}
}

\begin{document}
\maketitle
  
\begin{abstract}
We present AutoSchemaKG, a framework for fully autonomous knowledge graph construction that eliminates the need for predefined schemas. Our system leverages large language models to simultaneously extract knowledge triples and induce comprehensive schemas directly from text, modeling both entities and events while employing conceptualization to organize instances into semantic categories. Processing over 50 million documents, we construct ATLAS (Automated Triple Linking And Schema induction), a family of knowledge graphs with 900+ million nodes and 5.9 billion edges. This approach outperforms state-of-the-art baselines on multi-hop QA tasks and enhances LLM factuality. Notably, our schema induction achieves 95\% semantic alignment with human-crafted schemas with zero manual intervention, demonstrating that billion-scale knowledge graphs with dynamically induced schemas can effectively complement parametric knowledge in large language models\footnote{ \url{https://github.com/HKUST-KnowComp/AutoSchemaKG}}.
\end{abstract}

\section{Introduction}

In the current era of information abundance, transforming vast amounts of unstructured data into structured, machine-readable knowledge remains one of the most significant challenges in artificial intelligence.
Knowledge Graphs (KGs) have emerged as the cornerstone technology for this transformation \cite{zhao-etal-2024-divknowqa}, providing the semantic backbone for applications ranging from search engines and question answering \cite{wu-etal-2024-cotkr,chen-etal-2024-new,zong-etal-2024-triad,sun-etal-2024-head} to recommendation systems \cite{lyu-etal-2024-llm} and complex reasoning tasks \cite{li-etal-2024-assessing-logical}. Yet despite their critical importance, current KG construction approaches remain hampered by an inherent paradox: they require predefined schemas created by domain experts, which fundamentally limits their scalability, adaptability, and domain coverage.

\begin{figure*}[h]
    \centering
    \includegraphics[width=\linewidth]{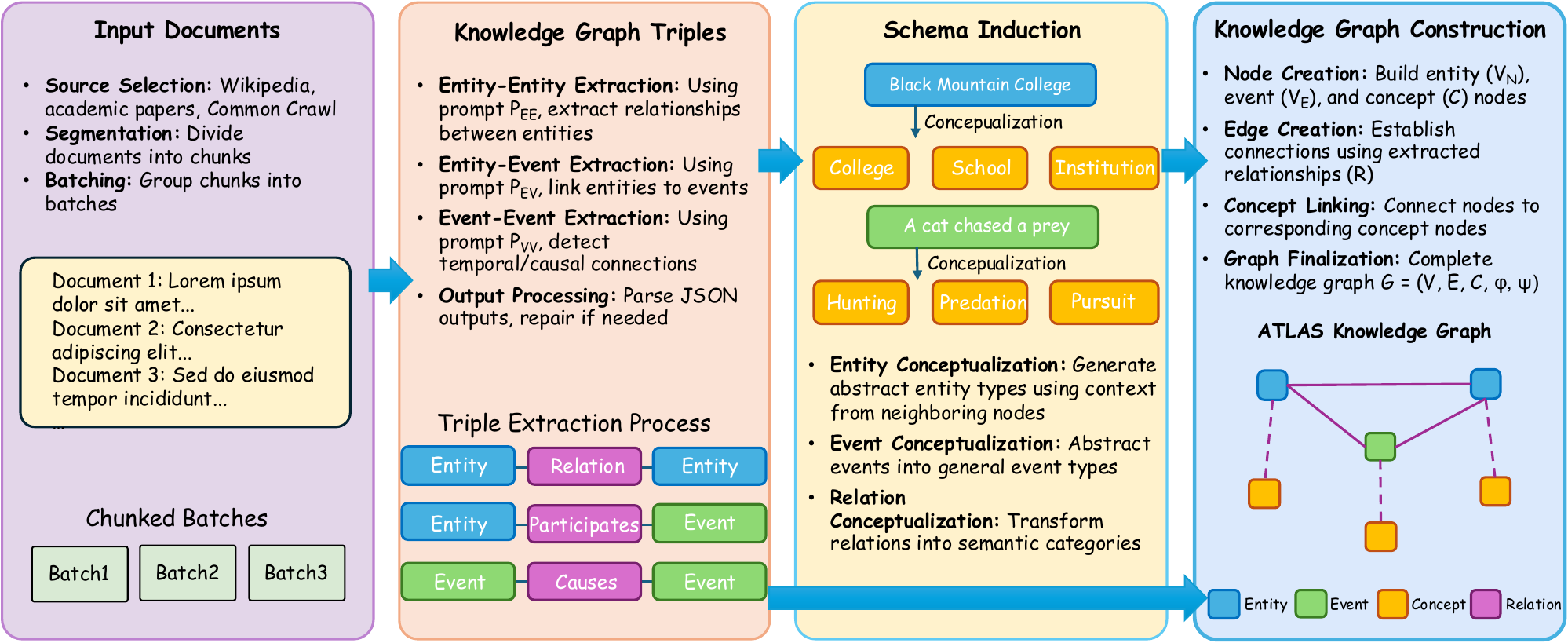}
    \vspace{-0.6cm}
    \caption{This figure illustrates the AutoSchemaKG pipeline for autonomous knowledge graph construction through four phases: (1) Input Processing: documents are filtered, segmented, and batched; (2) Triple Extraction: relationships between entities and events are extracted using LLM prompts; (3) Schema Induction: elements are conceptualized into abstract categories without predefined schemas; and (4) Knowledge Graph Construction: triples and schema are integrated into the ATLAS knowledge graph with entity nodes (blue), event nodes (green), concept nodes (orange), and relation edges (purple).}
    \vspace{-0.2cm}
    \label{fig:framework}
\end{figure*}

We present AutoSchemaKG, a paradigm-shifting framework that breaks this dependency by enabling autonomous knowledge graph construction without predefined schemas \cite{ye-etal-2023-schema}, as shown in Figure \ref{fig:framework}. Our approach leverages the semantic understanding capabilities of large language models to simultaneously extract knowledge triples and dynamically induce schemas directly from text, eliminating this manual bottleneck in KG development \cite{zhang-soh-2024-extract, li-etal-2024-integration,wang-etal-2025-llms-convert}. 

% Our framework distinguishes itself by modeling events alongside entities \cite{zhang2020aser, ZhangLPKOFS22}, recognizing that real-world knowledge is dynamic rather than static \cite{tan-etal-2024-set}. By treating events as  {\jx basic semantic units in the graph}, AutoSchemaKG captures temporal relationships, causality, and procedural knowledge missed by entity-only graphs. {\jx TODO: Add reference to Jacky works on discourse and strenthen the motivations. Also by adding text decomposition like DenseX Retrival. } Experiments confirm that with event information preserving over 90\% of original passage content versus just 70\% for entity information alone.

{\jx Our framework distinguishes itself by modeling events alongside entities \cite{zhang2020aser, ZhangLPKOFS22}, recognizing that real-world knowledge is dynamic rather than static \cite{tan-etal-2024-set}. By treating events as basic semantic units in the graph, AutoSchemaKG captures temporal relationships, causality, and procedural knowledge missed by entity-only graphs. This approach aligns with recent advancements in proposition-based text decomposition~\cite{hoyle-etal-2023-natural,jhamtani2024natural, chen-etal-2024-dense}, which demonstrate that by doing text composition, the resulting semantic units can provide better text representation from implicit meanings, meanwhile providing higher information density and retrieval accuracy than traditional passage-level representations. On the top of extracted events from text decomposition, our event-centric modeling is further supported by discourse relation recognition research \cite{chan-etal-2023-discoprompt, chan-etal-2024-exploring}, which emphasizes the importance of understanding implicit temporal and causal relationships between statements—connections that entity-only representations cannot adequately capture. This discourse-aware approach enables AutoSchemaKG to preserve hierarchical path information similar to the tree-structured discourse relations in natural text. Experiments confirm that with event information, AutoSchemaKG preserves over 90\% of original passage content versus just 70\% for entity information alone, demonstrating the crucial role of events in knowledge representation.}

Central to our innovation is the conceptualization \cite{wang-etal-2023-cat, bai2024intentionknowledgegraphconstruction,  wang-etal-2023-car, HE2024104149, wang-etal-2024-candle, wang-etal-2024-absinstruct} process that drives schema induction. Rather than simply extracting triples, we employ a sophisticated abstraction mechanism that generalizes specific entities, events, and relations into broader conceptual categories. This conceptualization serves multiple critical functions: it creates semantic bridges between seemingly disparate information, enables zero-shot inferencing across domains, reduces sparsity in KGs, and provides hierarchical organization that supports both specific and abstract reasoning \cite{wang-etal-2024-abspyramid, wang2024roleentityeventlevel}. Our approach transforms the traditional static schema into a dynamic, multi-level conceptual framework that adapts to new domains without predefined ontologies. This conceptualization layer represents a fundamental advancement, transforming a collection of disconnected facts into a coherent knowledge ecosystem with emergent reasoning capabilities.

By applying our framework to the Dolma 1.7 \cite{DBLP:journals/corr/abs-2402-00159} pretraining corpus across three diverse subsets, English Wikipedia, paper abstracts from Semantic Scholar, and 3\% of Common Crawl data, we construct the ATLAS family of knowledge graphs (ATLAS-Wiki, ATLAS-Pes2o, and ATLAS-CC). Collectively, these knowledge graphs comprise over 900 million nodes connected by 5.9 billion edges, containing billions of facts that are comparable in scale to the parametric knowledge stored in large language models. This scale is crucial, as we demonstrate that knowledge graphs must reach a critical mass of billions of facts to effectively compete with and complement the parametric knowledge in contemporary LLMs.

Our work addresses a fundamental research question: can retrieval-augmented generation with comparable-sized knowledge graphs enhance LLM performance even when the retrieval corpus is drawn from the same sources as the model's pretraining data? Through comprehensive evaluations on general benchmarks, we demonstrate that properly structured knowledge representations offer advantages over traditional text-based retrieval even in these scenarios. Most significantly, our schema induction process achieves 92\% semantic alignment with human-crafted schemas with zero manual intervention, demonstrating that automated schema generation can match expert quality while dramatically improving construction efficiency.

The power of AutoSchemaKG extends beyond its construction methodology to deliver substantial performance improvements in downstream applications. In rigorous evaluations, our approach outperforms state-of-the-art baselines by 12-18\% on multi-hop question answering tasks \cite{trivedi2021musique,yang2018hotpotqa,ho-etal-2020-constructing} and enhances large language model factuality by up to 9\% \cite{chen2023felm}. Moreover, we found that our constructed knowledge graph is helpful for Llama3.1 7B models on general reasoning task on various domains that intensively requeries background knowledge, including Global Facts, History, Law, Religion, Philosophy and Ethics, Medicine and Health, and Social Sciences. 
These gains stem from our system's ability to create richer semantic representations through the integration of entities, events, and their conceptual abstractions—enabling better reasoning over complex information across different domains and data sources.
Our key contributions include:
\begin{itemize}
    \item We develop an entity-event-concept extraction framework that captures not only traditional entity relationships but also complex event structures and their conceptual categorizations, creating a multi-dimensional knowledge representation.
    \item We apply our efficient knowledge extraction and integration approach to web-scale data, processing billions of triples while maintaining semantic consistency. The resulting ATLAS-family knowledge graphs are, to the best of our knowledge, both the largest automatically constructed knowledge graphs and the largest graph-based Retrieval Augmented Generation (Graph RAG) datasets available.
    \item We build a retrieval augmented generation pipeline on the billion-scale ATLAS KGs, demonstrating AutoSchemaKG's effectiveness across diverse domains without domain-specific customization, establishing a truly general-purpose knowledge acquisition framework.
\end{itemize}

AutoSchemaKG represents a fundamental rethinking of knowledge graph construction, transforming what was once a heavily supervised process requiring significant domain expertise into a fully automated pipeline. This advancement not only accelerates KG development but also dramatically expands the potential application domains for knowledge-intensive AI systems.

\section{Problem Definition}

We formally outline the tasks involved in automatically constructing knowledge graphs. We begin by providing a precise definition of a knowledge graph equipped with a conceptual schema.
\begin{definition}[Knowledge Graph with Conceptual Schema]
Consider a knowledge graph denoted as $G = (V, E, C, \phi, \psi)$, where:
$V = V_E \cup V_N$ represents the collection of nodes, with $V_E$ as the set of event nodes, $V_N$ as the set of entity nodes, and $V_E \cap V_N = \emptyset$.
$E \subseteq V \times V \times R$ defines the set of edges, where $R$ denotes relation types. Edges may connect entity-entity, entity-event, or event-event nodes.
$C$ is the set of conceptual categories.
$\phi: V \rightarrow \mathcal{P}(C)$ assigns each node a subset of concepts, where $\phi(v) \subseteq C$ for every $v \in V$.
$\psi: R \rightarrow \mathcal{P}(C)$ links each relation type to a subset of concepts, where $\psi(r) \subseteq C$ for every $r \in R$.
$\mathcal{P}(C)$ denotes the power set of $C$, encompassing all possible subsets.
Additional constraints: $\forall v \in V: \phi(v) \neq \emptyset$ and $\forall r \in R: \psi(r) \neq \emptyset$.
\end{definition}

\section{AutoSchemaKG Framework}

In this section, we elaborate on the process of fully automating knowledge graph construction. 
% Given a collection of $n$ documents $D = \{D_1, D_2, D_3, \dots, D_n\}$, our method systematically builds the graph. 

\begin{table*}[]
\centering
\small
\begin{tabular}{@{}l|rrr|rrr@{}}
\toprule
 & \multicolumn{3}{c|}{Question Answering Corpora} & \multicolumn{3}{c}{Pre-training Corpora} \\ \midrule
 & \multicolumn{1}{c}{MuSiQue} & \multicolumn{1}{c}{2WikiQA} & \multicolumn{1}{c|}{HotpotQA} & \multicolumn{1}{c}{ATLAS-Wiki} & \multicolumn{1}{c}{ATLAS-Pes2o} & \multicolumn{1}{c}{ATLAS-CC} \\ \midrule
\# Text Chunks & 11,656 & 6,119 & 9,221 & 9.599M & 7.918M & 35.040M \\
\# Entities & 108,582 & 48,782 & 95,686 & 70.104M & 75.857M & 241.061M \\
\# Events & 99,747 & 50,910 & 82,833 & 165.717M & 92.636M & 696.195M \\
\# Concepts & 37,414 & 19,830 & 32,410 & 8.091M & 5.895M & 31.070M
\\
\# Nodes & \textbf{245,743} & \textbf{119,522} & \textbf{210,929} & \textbf{243.912M} & \textbf{174.387M} & \textbf{937.256M} \\
\midrule

\# Entity-Entity Edges & 91,186 & 40,748 & 78,467 & 0.114B & 0.076B & 0.414B \\
\# Event-Entity Edges & 143,254 & 63,680 & 123,527 & 0.265B & 0.208B & 1.063B \\
\# Event-Event Edges & 45,157 & 21,062 & 36,602 & 0.071B & 0.044B & 0.295B \\
\# Conceptulization Edges & 933,330 & 432,869 & 789,608 & 1.041B & 0.821B & 4.178B \\
\# Edges & \textbf{1,212,927} & \textbf{558,359} & \textbf{1,028,204} & \textbf{1.492B} & \textbf{1.150B} & \textbf{5.958B} \\
\bottomrule
\end{tabular}
\vspace{-0.3cm}
\caption{Statistics of knowledge graph construction across QA datasets (MuSiQue, 2WikiQA, HotpotQA) and LLM pre-training corpora (En-Wiki, Pes2o-Abstract, Common Crawl) for ATLAS knowledge graphs, showing counts of text chunks, nodes (entities/events), concepts, edges (Entity-Entity, Event-Entity, Event-Event), and conceptualizations. M = million, B = billion.}
\vspace{-0.3cm}
\label{tab:kg_statistics}
\end{table*}

\subsection{Triple Extraction}
\label{sec:triple_extraction}
Our approach employs a multi-phase pipeline using Large Language Models to convert unstructured text into knowledge triples from the Dolma corpus \cite{DBLP:journals/corr/abs-2402-00159}. This pipeline extracts Entity-Entity, Entity-Event, and Event-Event relationships through three sequential stages.
We preprocess texts by filtering for English language and segmenting documents that exceed token limits. The segmented texts are grouped into processing batches.
Stage 1 extracts Entity-Entity relationships using a system prompt $P_{EE}$ that instructs the LLM to detect entities and their interrelations. The output is parsed into triples $(e_1, r, e_2)$, where $e_1, e_2 \in V_N$ are entity nodes and $r \in R$ is a relation type.
Stage 2 identifies Entity-Event relationships with prompt $P_{EV}$, producing triples $(e, r, v)$ or $(v, r, e)$, where $e \in V_N$, $v \in V_E$, and $r \in R$.
Stage 3 targets Event-Event relationships with prompt $P_{VV}$, generating triples $(v_1, r, v_2)$, where $v_1, v_2 \in V_E$ and $r \in R$.
The pipeline supports various LLMs with optimized precision settings and GPU acceleration. Extracted triples with their corresponding texts and metadata are serialized into JSON files.

\subsection{Schema Induction}
\label{sec:schema_induction}

Following triple extraction, we perform schema induction to abstract specific entities, events, and relations into generalized types. This process uses LLMs to generate conceptual phrases representing types of each graph element, aligning with our formal definition $G = (V, E, C, \phi, \psi)$.
For each category (events, entities, and relations), we process elements in batches. The LLM generates at least three phrases per element that encapsulate its type or related concepts at varying abstraction levels.
For entities ($e \in V_N$), we enhance abstraction by incorporating contextual information from neighboring nodes. We sample up to $N_{ctx}$ neighbors to construct a context string that provides additional semantic cues.
The schema induction pipeline processes the graph serialized from the triple extraction phase. Elements are partitioned into batches, with options for slicing for distributed computation. The generated phrases are recorded in a CSV file, mapping each node $v \in V$ and relation $r \in R$ to a subset of concepts in $C$ via $\phi$ and $\psi$.
This automated schema enhances the knowledge graph's adaptability across varied domains without requiring manual curation.

\section{Construction of ATLAS Families}

\paragraph{Corpora} 
As shown in Table \ref{tab:kg_statistics}, the ATLAS-Wiki, ATLAS-Pes2o, and ATLAS-CC are constructed from the subsets from Dolma's subset of Wikipedia \& Wikibooks, Semantic Scholar, and Dolma's CC respectively.\footnote{\url{https://huggingface.co/datasets/allenai/dolma}} We use the full Wikipedia \& Wikibooks to construct the ATLAS-Wiki, and we use the abstract part of Semantic Scholar to construct ATLAS-Pes2o, and we use the each of 3\% from cc-head, cc-middle, and cc-tail to construct ATLAS-CC. According to \cite{DBLP:journals/corr/abs-2402-00159} the head, middle, tail of CC are used to measure the distribution similarity to Wikipedia text.

\paragraph{Computational Cost}
We constructed our knowledge graphs using 80GB GPUs with 1,513 TFLOPS of FP16 compute, running \texttt{Llama-3-8B-instruct} with Flash Attention. The computational demands were substantial: 14,300 GPU hours for En-Wiki (243.9M nodes, 1.49B edges), 11,800 GPU hours for Pes2o-Abstract (174.4M nodes, 1.15B edges), and 52,300 GPU hours for Common Crawl (937.3M nodes, 5.96B edges). Processing 1024-token chunks in batches, we invested approximately 78,400 GPU hours total to extract billions of semantic relationships.

\section{Experiment}

In the this section, we show that the AutoSchemaKG has accurate triplex extraction, can coherently induce schemas, and has very high information preservations in section \ref{subsec:quality_of_framework}.

\begin{table}[ht]
\centering
\small
\vspace*{-0.25cm}
\setlength\tabcolsep{4pt}%调列距
\renewcommand{\arraystretch}{1.2}%调行距
 \resizebox{0.48\textwidth}{!}{\begin{tabular}{l|c|ccc}
\toprule
\multirow{1}{*}{\textbf{Knowledge Graph}} &
\multirow{1}{*}{\textbf{Triple Type}}&
\textbf{Precision} & \textbf{Recall} & \textbf{F1} \\ 
\midrule    
\multirow{3}{*}{ATLAS-Wiki}&Entity-Entity& 99.13& 90.10& 94.09\\
                            & Event-Entity& 100.0& 92.59& 95.60\\
                            & Event-Event& 99.60& 93.59&96.01\\
    \midrule
\multirow{3}{*}{ATLAS-Pes2o}&Entity-Entity& 97.66& 89.89& 93.03\\
                            &Event-Entity& 100.0& 94.29& 96.83\\
                            &Event-Event& 99.54& 91.31& 94.94\\
    \midrule
\multirow{3}{*}{ATLAS-CC}&Entity-Entity& 95.65& 84.64& 88.82\\
                            &Event-Entity& 99.93& 87.92& 92.72\\
                            &Event-Event& 99.86& 93.20& 96.16\\
\bottomrule
\end{tabular}
}
\vspace{-0.3cm}
\caption{Triple precision, recall and F1 score across datasets. Each row displays the performance of a type of extracted triples.}
\label{tab:triple_result}
\vskip -0.1in
\end{table}

\subsection{Evaluating AutoSchemaKG}

% \subsection{Triple Precision} 
\label{subsec:quality_of_framework}
 
\paragraph{Evaluating Triple Extraction Accuracy} We use a rigorous counting-based evaluation method. Rather than relying on subjective scoring, we employ DeepSeek-V3~\cite{liu2024deepseek} as a judge in a structured verification process. For each document: (1) We present DeepSeek-V3 with both the original text and the triples extracted by Llama-3-8B-Instruct;
(2) DeepSeek-V3 identifies triples that are incorrectly extracted (false positives);
(3) DeepSeek-V3 lists facts present in the original text but missing from the extracted triples (false negatives);
This methodology allows us to calculate precise metrics:
(1) Precision: proportion of correctly extracted triples out of all extracted triples;
(2) Recall: proportion of correctly extracted triples among all ground-truth triples in the text;
(3) F1 score: the harmonic mean of precision and recall.
As shown in Table ~\ref{tab:triple_result}, our approach demonstrates exceptional extraction quality across all datasets, with particularly strong performance on Wikipedia content. The precision, recall, and F1 scores of the triples in our KG exceed 90\% in most cases, demonstrating the high quality and reliability of our extracted triples.

\begin{table}[ht]
\centering
\small
% \vspace*{-0.25cm}
\setlength\tabcolsep{4pt}%调列距
\renewcommand{\arraystretch}{1.2}%调行距
 \resizebox{0.48\textwidth}{!}{\begin{tabular}{l|c|cc}
\toprule
\multirow{2}{*}{\textbf{Dataset}} &
\multirow{2}{*}{\textbf{Context}} &
\multicolumn{2}{c}{\textbf{Model}} \\
\cmidrule(lr){3-4}
     & & \textbf{LLaMA-3-8B} & \textbf{LLaMA-3-70B} \\ \midrule
    
\multirow{4}{*}{ATLAS-Wiki}& [Lower-Upper]& 46.29-99.29& 65.69-99.70\\
                            & Entity& 65.08& 70.96\\
                            & Event& \underline{92.69}& \underline{94.82}\\
                            & Event + Entity& \textbf{93.30}& \textbf{95.13}\\
    \midrule
\multirow{4}{*}{ATLAS-Pes2o}& [Lower-Upper]& 62.32-98.99& 75.05-99.49\\
                            & Entity& 80.00& 83.33\\
                            & Event& \underline{96.97}& \underline{97.78}\\
                            & Event + Entity& \textbf{97.37}& \textbf{97.98}\\
    \midrule
\multirow{4}{*}{ATLAS-CC}& [Lower-Upper]& 56.08-97.29& 70.25-99.10\\
                            & Entity& 76.78& 81.01\\
                            & Event& \underline{94.87}& \underline{96.78}\\
                            & Event + Entity& \textbf{96.28}& \textbf{96.98}\\
\bottomrule
\end{tabular}
}
\vspace{-0.3cm}
\caption{KG performance across datasets, showing bounds (no context to full passage) and results with different knowledge representations. Entity, Event, and combined representations preserve most information for MCQs, approaching full-passage performance across all datasets and models.}
\label{tab:mcq_result}
\vspace{-0.5cm}
\end{table}

\paragraph{Measuring Information Preservation in Knowledge Graphs} We evaluate the effectiveness of the entity-level triples and event-level triples of our constructed KG in preserving information from original passages. We test how well {multiple-choice question} { (MCQ)} performance is preserved when we convert the original passage into KG data.
Following the evaluation protocol from the existing work~\cite{schuhmann2025project}, we generate five MCQs with \texttt{LLaMA-3-70B-Instruct} for each original passage, and the prompts are shown in Figure \ref{fig:mcq_prompts}. We sample 200 original passages and 1,000 MCQs are obtained for each dataset. We ask LLMs to answer them with no context (denoted as lower bound), then ask them again with the original passage (denoted as upper bound) for sanity check. Finally, we conduct tests using entity-level triples (denoted as Entity), event-level triples (denoted as Event), and a combination of both entity-level and event-level triples (denoted as Event + Entity). We evaluate on three pre-training datasets with our constructed KG in Table~\ref{tab:kg_statistics}: En-Wiki, Pes2o-Abstract and Common Crawl. According to the results shown in Table~\ref{tab:mcq_result}, we have the following insights:
(1) \textbf{Information is well preserved in our constructed KG.} MCQs performance with Entity, Event or Event + Entity remains far above the lower bound baseline and approaches the original-passage upper bound. It suggests that the information in the original passages is well preserved in our constructed KG;
(2) \textbf{Events are more effective than entities.} The MCQs performance with Event or Event + Entity is much closer to the upper bound than that with Entity, which accuracy is more than 95\% in most of the cases. It demonstrates that the event-level triples can preserve richer and more precious information than entity-level triples.

\begin{table}[ht]

\setlength\tabcolsep{4pt}%调列距
\renewcommand{\arraystretch}{1.2}%调行距
\centering
 \resizebox{0.48\textwidth}{!}{\begin{tabular}{l|l|l|cc}
\toprule
     & \textbf{Task}& \textbf{Dataset} & \textbf{BS-R}&\textbf{BS-C}\\ 
\midrule
\multirow{4}{*}{LLaMA-3-8B} & \multirow{2}{*}{\textit{Entity Typing}}& FB15kET& 88.57& 86.54\\
                            &                              & YAGO43kET& 80.67& 58.86\\
                            & \multirow{1}{*}{\textit{Event Typing}}& wikiHow& 99.18& 99.26\\
                            & \multirow{1}{*}{\textit{Relation Typing}}& FB15kET& 88.75& 88.41\\
    \midrule
\multirow{4}{*}{LLaMA-3-13B} & \multirow{2}{*}{\textit{Entity Typing}}& FB15kET& 89.25& 88.59\\
                            &                              & YAGO43kET& 94.26& 90.56\\
                            & \multirow{1}{*}{\textit{Event Typing}}& wikiHow& 98.97& 99.33\\
                            & \multirow{1}{*}{\textit{Relation Typing}}& FB15kET& 88.58& 88.66\\
    \midrule
\multirow{4}{*}{LLaMA-3-70B} & \multirow{2}{*}{\textit{Entity Typing}}& FB15kET& 89.49& 87.30\\
                            &                              & YAGO43kET& 94.61& 92.64\\
                            & \multirow{1}{*}{\textit{Event Typing}}& wikiHow& 99.41& 99.15\\
                            & \multirow{1}{*}{\textit{Relation Typing}}& FB15kET& 88.70& 90.33\\
    \bottomrule
\end{tabular}
}
\vspace{-0.3cm}
\caption{Results of schema induction with various LLaMA family LLMs across three kinds of typing tasks. }
\label{tab:result_schema_induction}
\vspace{-0.3cm}
\end{table}

\paragraph{Measuring Schema Quality}  To demonstrate our schema induction method's capability, we apply it to entity, event, and relation typing tasks, measuring how many types our method recalls. Dataset details are in Appendix~\ref{appendix:sa_dataset}. Since rule-based evaluation might overlook semantic similarities, we use two semantic-level metrics: \textbf{BS-R} and \textbf{BS-C}, explained in Appendix~\ref{appendix:sa_metric}.
Table~\ref{tab:result_schema_induction} shows results using three sizes of LLaMA-3. Our method achieves over 80\% and often 90\% recall for entity, event, and relation types in most cases, with performance improving as LLM parameter size increases.

\subsection{Performance on Multi-hop QA Tasks}
This subsection details the experimental setup for open-domain QA, focusing on multi-hop reasoning tasks where our knowledge graph's structure and schema induction are expected to excel.

\paragraph{Datasets}
We select three benchmark datasets known for their multi-hop reasoning demands, all derived from Wikipedia: MuSiQue \cite{trivedi2021musique}, HotpotQA \cite{yang2018hotpotqa}, and 2WikiMultihopQA \cite{ho-etal-2020-constructing}, necessitating complex relational paths across articles. From each dataset, we randomly select one thousand questions following \cite{gutiérrez2024hipporag}.

\paragraph{Baselines and Metrics}
We compare our knowledge graph-based RAG system against several state-of-the-art approaches. The graph-based baselines include HippoRAG \cite{gutiérrez2024hipporag}, a framework that builds a memory graph from text using entity recognition and relation extraction; HippoRAG2 \cite{gutierrez2025rag}, an advanced iteration with enhanced graph construction; GraphRAG \cite{edge2024local}, Microsoft Research's technique combining text extraction, network analysis, and LLM prompting; LightRAG \cite{guo2024lightrag}, a simpler alternative to GraphRAG focused on efficiency; and MiniRAG \cite{fan2025minirag}, an extremely simple framework tailored for Small Language Models. For MiniRAG and LightRAG we adjust its prompts so that it outputs brief answers for QA instead of generating long answers with explanations. 
For text-based RAG comparisons, we evaluate against BM25 + LLM using traditional retrieval with BM25 scoring; Contriever \cite{izacard2021contriever}, a dense retrieval-augmented system fine-tuned for QA; 
% ColBERTv2 \cite{santhanam-etal-2022-colbertv2}, a late-interaction model for efficient retrieval; 
and RAPTOR \cite{sarthi2024raptor}, a hierarchical summarization system. These baselines allow us to benchmark our approach against diverse retrieval-augmented methods.

We evaluate our system using standard metrics for open-domain QA. Exact Match (EM) measures binary correctness after normalization: $\text{EM}(a, g) = \mathbf{1}[\text{norm}(a) = \text{norm}(g)]$, where $a$ is the predicted answer, $g$ is the gold answer, and normalization includes lowercasing and removing articles, punctuation, and whitespace. F1 Score measures token overlap between normalized answers: $\text{F1} = \frac{2 \cdot P \cdot R}{P + R}$, where $P = |a \cap g|/|a|$ and $R = |a \cap g|/|g|$ are precision and recall.

\begin{table}[t]
\scriptsize
\centering
\resizebox{\linewidth}{!}{
\begin{tabular}{@{}l|cc|cc|cc@{}}
\toprule
\textbf{Model/Dataset} & \multicolumn{2}{c|}{\textbf{MuSiQue}} & \multicolumn{2}{c|}{\textbf{2Wiki}} & \multicolumn{2}{c}{\textbf{HotpotQA}} \\ \midrule
\textbf{Metric} & \textbf{EM} & \textbf{F1} & \textbf{EM} & \textbf{F1} & \textbf{EM} & \textbf{F1} \\ \midrule
\rowcolor[gray]{0.9} \multicolumn{7}{l}{\textit{Baseline Retrievers}} \\ \midrule
No Retriever & 17.6 & 26.1 & 36.5 & 42.8 & 37.0 & 47.3 \\
Contriever & 24.0 & 31.3 & 38.1 & 41.9 & 51.3 & 62.3 \\
BM25 & 20.3 & 28.8 & 47.9 & 51.2 & 52.0 & 63.4 \\ \midrule

\rowcolor[gray]{0.9} \multicolumn{7}{l}{\textit{LLM Embeddings}} \\ \midrule
GTE-Qwen2-7B& 30.6 & 40.9 & 55.1 & 60.0 & 58.6 & 71.0  \\
GritLM-7B & 33.6 & 44.8 & 55.8 & 60.6 & 60.7 & 73.3  \\
NV-Embed-v2 (7B) & \underline{34.7} & {45.7} & 57.5 & 61.5 & \textbf{62.8} & {75.3} \\ \midrule

\rowcolor[gray]{0.9} \multicolumn{7}{l}{\textit{Existing Graph-based RAG Methods}} \\ \midrule
RAPTOR & 20.7 & 28.9 & 47.3 & 52.1 & 56.8 & 69.5 \\
GraphRAG & 27.3 & 38.5 & 51.4 & 58.6 & 55.2 & 68.6 \\
LightRAG  & 20.0 & 29.3 & 38.6 & 44.6 & 33.3 & 44.9 \\
MiniRAG  & 9.6 & 16.8 & 13.2 & 21.4 & 47.1 & 59.9 \\ 
HippoRAG & 26.2 & 35.1 & 65.0 & {71.8} & 52.6 & 63.5 \\
HippoRAG2 & \textbf{37.2} & \textbf{48.6} & {65.0} & 71.0 & \underline{62.7} & {75.5} \\ 
\midrule\midrule
\rowcolor[gray]{0.9} \multicolumn{7}{l}{\textit{AutoSchemaKG (\texttt{LLama-3.1-8B-Instruct}) + Think-on-Graph}} \\ \midrule
Entity-KG & 14.8 & 26.0 &36.9 & 44.0 & 41.9 &55.2  \\
Entity-Event-KG  & 19.4 & 32.8 &39.0 & 47.1 & 47.7 &61.2 \\
Full-KG  &  20.1 & 31.2 &40.0 & 47.7 & 48.2 &60.5 \\ \midrule
\rowcolor[gray]{0.9} \multicolumn{7}{l}{\textit{AutoSchemaKG (\texttt{LLama-3.1-8B-Instruct}) + HippoRAG}} \\ \midrule
Entity-KG  & 22.5 & 36.4 & {57.7} & {65.8} & 50.3 & 65.8 \\
Entity-Event-KG  & {22.9} & {36.1} & 56.4 & 64.4 & {48.6} & {64.6} \\
Full-KG  & 23.6 & 36.5 & 54.8 & 63.2 & 50.0 & 65.3 \\ \midrule
\rowcolor[gray]{0.9} \multicolumn{7}{l}{\textit{AutoSchemaKG (\texttt{LLama-3.1-8B-Instruct}) + HippoRAG2}} \\ \midrule
Entity-KG  & 31.4 & 47.2 & 64.2 & 73.3 & 60.9 & \underline{77.5} \\
Entity-Event-KG & 31.6 & 47.3 & \underline{65.2} & \underline{73.7} & 60.0 & 77.0 \\
Full-KG  & {31.8} & \underline{47.3} & \textbf{65.3} & \textbf{73.9} & {61.8} & \textbf{78.3} \\ \bottomrule
\end{tabular}
}
\vspace{-0.2cm}
\caption{Performance comparison of AutoSchemaKG integrated with ToG, HippoRAG and HippoRAG2 with bold indicating the highest performance per dataset.}
\label{tab:qa_performance}\vspace{-0.2cm}
\end{table}

\paragraph{Think on Graph Settings}
We implement Think on Graph (ToG) \cite{sun2024thinkongraph} using a knowledge graph derived from our corpus. Nodes represent extracted entities and concepts, with edges showing semantic relations. We use \texttt{multi-qa-MiniLM-L6-dot-v1} to compute embeddings for all graph elements, indexed with FAISS. \texttt{LLama-3.3-70B-Instruct} performs entity recognition, path scoring, reasoning, and answer generation. The workflow extracts query entities, retrieves starting nodes, explores graph paths through depth-limited search, prunes irrelevant paths, and generates answers based on retrieved paths. Algorithm \ref{alg:tog} in the Appendix provides details.

\paragraph{HippoRAG 1\&2 Settings}
In our implementation of HippoRAG \cite{gutiérrez2024hipporag}, we extend the original framework to operate on a customized graph. Initially presented in the foundational paper, we employ Named Entity Recognition (NER) to build a personalized dictionary for PageRank execution.
Regarding HippoRAG2 \cite{gutierrez2025rag}, we select the top 30 edges (musique dataset 50 edges) for LLM filtering, incorporating a weight adjustment factor of 0.9. Considering the capability of our graph to effectively locating subgraphs, combined with various graph configurations (entity, event, concept) resulting in graphs of differing densities, we set the damping factor to 0.9 to concentrate on propagation within the local subgraph. For further implementation details, please refer to Algorithm \ref{fig:hipporag2_algo}.

\paragraph{Implementation Details}
The knowledge graph is constructed from the corresponding context corpora for each dataset folowing \cite{gutiérrez2024hipporag} using the framework of AutoSchemaKG with $L_{max} = 1024$ and $B = 16$, and the schema induction pipeline (Section~\ref{sec:schema_induction}) with $B_s = 5$. We employ Meta’s \texttt{LLaMA-3.1-8B-Instruct} to construct the graphs, optimized with bfloat16 precision and Flash Attention 2. The graph is stored in NetworkX for retrieval, with subgraphs fed into \texttt{LLaMA-3.3-70B-Instruct} for answer generation. 

\paragraph{Evaluation Results}
The experimental results in Figure \ref{tab:qa_performance} and Figure \ref{tab:recall_performance} demonstrate AutoSchemaKG's effectiveness in multi-hop question answering across three benchmark datasets. With HippoRAG2 integration, the Full-KG configuration (entities, events, and concepts) outperforms traditional retrieval approaches like BM25 and Contriever by 12-18\%, highlighting its strength in complex reasoning scenarios. Notably, AutoSchemaKG achieves comparable or better results using \texttt{LLaMA-3.1-8B-Instruct} as graph constructor compared to the original HippoRAG2 implementation that requires \texttt{LLaMA-3.3-70B-Instruct} for both construction and QA reading.

\begin{table}[t]
\centering
\small
\begin{tabular}{l|l|c c}
\toprule
\textbf{Corpus} & \textbf{Method} & \textbf{Acc} & \textbf{F1} \\
\midrule
- & - & 54.08 & 26.79 \\
\midrule
\rowcolor[gray]{0.9} \multicolumn{4}{l}{\textit{Text Corpora}}\\
\midrule

\multirow{3}{*}{Wikipedia} & Random & 52.77 & 25.56 \\
& BM25 & \underline{56.15} & \underline{30.43} \\
& Dense Retrieval & 56.04 & 30.33 \\
\midrule

\multirow{3}{*}{Pes2o-Abstract} & Random & 53.34 & 26.00 \\
& BM25 & 54.60 & 27.95 \\
& Dense Retrieval & 55.43 & 29.19 \\
\midrule

\multirow{3}{*}{Common Crawl} & Random & 53.31 & 26.45 \\
& BM25 & 54.56 & 28.32 \\
& Dense Retrieval & 54.42 & 28.49 \\
\midrule
\rowcolor[gray]{0.9} \multicolumn{4}{l}{\textit{Knowledge Graph}}\\
\midrule
Freebase & Think on Graph & 53.75 & 24.81 \\
\midrule
ATLAS-Wiki &  & \textbf{56.43} & \textbf{30.48} \\
% \cmidrule{1-1} \cmidrule{3-4}
ATLAS-Pes2o& HippoRAG2 & 55.30 & 28.12 \\
ATLAS-CC &  & 55.56 & 29.57 \\
% \cmidrule{1-1} \cmidrule{3-4}

\bottomrule
\end{tabular}
\vspace{-0.2cm}
\caption{Balanced accuracy (\%) and F1 score (\%) on FELM benchmark of \texttt{Llama-3.1-8b-Instruct} with retrieval methods. The best results are in \textbf{bold}, and the second best results are \underline{underlined}.}
\label{tab:fact_results}
\vspace{-0.2cm}
\end{table}

\paragraph{Advantages of Events and Concepts}
Our case studies revealed two key benefits of event and concept nodes:
\textbf{1) Event nodes provide enriched context.} As shown in Figure \ref{fig:event-example}, they serve as valuable retrieval targets when critical information in triples is ambiguous or missed, helping identify relevant subgraphs containing passage nodes;
\textbf{2) Concept nodes create alternative pathways.} These nodes establish connections beyond direct entities and events, addressing complex multi-hop question answering limitations. Figure \ref{fig:concept-example} shows how concept nodes link knowledge across disparate subgraphs, enabling systems like HippoRAG to bridge separate subgraph influences via PageRank algorithms.

\begin{table*}[t]
% \huge
\small
\centering
% \resizebox{0.75\linewidth}{!}{
\begin{tabular}{@{}l|cccccccc|c@{}}
\toprule
Knowledge Source & \multicolumn{1}{l}{History} & \multicolumn{1}{l}{Law} & \multicolumn{1}{l}{Religion} & \multicolumn{1}{l}{Phil/Eth} & \multicolumn{1}{l}{Med/Hlth} & \multicolumn{1}{l}{GlbFct} & \multicolumn{1}{l}{SocSci} & \multicolumn{1}{l}{Logic}& \multicolumn{1}{l}{Average} \\ \midrule
 None &76.59 &66.86&83.04& 63.55& 70.38& 66.72 & 79.74& 64.35&72.10\\
 Freebase-ToG &\textbf{78.42} &69.00&75.44& \underline{65.67}& \underline{72.65}& 67.27 & 76.00& 66.03&72.34\\
\midrule
\rowcolor[gray]{0.9} \multicolumn{10}{l}{\textit{Random Baseline}} \\
\midrule
Wikipedia& 76.64& 66.82&79.53& 59.26& 70.34& 66.46& 77.78&59.21 & 70.62\\
 Common Crawl &74.89 & 66.52&79.53& 61.74& 69.82& 68.11 & 77.52&59.30&70.60\\
 Pes2o-Abstract & 76.24& 64.16&80.70 &62.01 & 70.62& 66.59 & 77.16&62.39&70.82\\
\midrule
\rowcolor[gray]{0.9} \multicolumn{10}{l}{\textit{Text Corpora + DBM25}} \\
\midrule
 Wikipedia &76.67 & 67.35&78.36&63.34 &69.35 & 61.98&76.99& 62.30&70.61\\
 Common Crawl&76.15 & 66.36& 80.12& 60.43 & 69.58& 64.67&76.71&63.18& 70.36\\
Pes2o-Abstract &78.01& 65.89 & 78.95&63.83 &71.01 & 65.78 &77.07 &59.34 &71.22\\
\midrule
\rowcolor[gray]{0.9} \multicolumn{10}{l}{\textit{Text Corpora + Dense Retrieval}} \\
\midrule
Wikipedia &73.59 & \underline{69.60}&79.53 & 63.58& 70.82& 62.41 &76.83& 62.21&70.86\\
Common Crawl &74.47 &68.98 &79.53 &60.46 &69.29 & 64.09 & 75.21&61.86& 69.56\\ 
Pes2o-Abstract&75.79 &61.82 & 78.36& 65.15& 69.72& 66.77& 76.47 & 63.05 &70.52 \\  
\midrule
\rowcolor[gray]{0.9} \multicolumn{10}{l}{\textit{ATLAS + HippoRAG2 }} \\
\midrule
ATLAS-Wiki & 76.73& 67.38& \underline{84.21}& \textbf{66.01}& 70.82& \underline{68.36} &79.16&63.65 &72.53\\ 
ATLAS-CC & \underline{78.16}& \textbf{70.85}& 83.04&65.60& 71.28&  63.95&  78.16&65.42&72.66\\
ATLAS-Pes2o& 77.13&68.41& 81.29& 65.05&\textbf{72.75}  & 65.67  & \underline{81.19}& 62.98&\underline{73.25}\\ 
\midrule
\rowcolor[gray]{0.9} \multicolumn{10}{l}{\textit{ATLAS + ToG}} \\
\midrule
ATLAS-Wiki  & 77.91&66.60&\underline{84.21}& 65.10& 70.69& 63.85&78.31&\underline{67.08}&72.18\\ 
ATLAS-CC  &77.07&68.18&83.63&65.24& 72.03& 66.87&79.72& 66.59 & 73.07\\
ATLAS-Pes2o&77.52&66.95&\textbf{84.80}& 63.44&71.15&\textbf{68.92}&\textbf{81.59}&\textbf{67.87}& \textbf{73.28}\\ 
\bottomrule
\end{tabular}
% }
% \vspace{-0.3cm}
\caption{Performance comparison of \texttt{Llama-3.1-8b-Instruct} with our KG-integrated HippoRAG2 and ToG versus baseline methods across Wikipedia, Common Crawl, and Pes2o-Abstract corpora on MMLU benchmarks. Tasks are grouped by subject, with bold and underlined values indicating first and second-highest scores. Phil/Eth, Med/Hlth, GlbFct, and SocSci denote Philosophy/Ethics, Medicine/Health, Global Facts, and Social Sciences.}
% \vspace{-0.3cm}
\label{tab:general_benchmarks}
\end{table*}

\subsection{Enhancing LLM Factuality with KGs}

We evaluated our KG's effectiveness in enhancing factuality using the FELM benchmark~\cite{chen2023felm}, which contains 847 samples across five domains with 4,425 fine-grained text segments. Following FELM's protocol, we applied RAG to three domains (world knowledge, science/technology, and writing/recommendation) while maintaining vanilla settings for math and reasoning domains.
For a comprehensive comparison, we evaluated against multiple retrieval methods: HippoRAG v2, BM25, and dense retrieval using MiniLM~\cite{minilmv2}. The retrieval process on text corpora is implemented in ElasticSearch database system~\cite{elasticsearch2018elasticsearch}. Our decision to use MiniLM rather than larger language model-based embedding approaches was driven by computational constraints. Implementing dense retrieval with higher-dimensional embeddings (e.g., 4096 dimensions) across our one billion nodes would require approximately 16 terabytes of storage using standard 32-bit floating-point representation.
These baselines represent state-of-the-art approaches in graph-based RAG and standalone retrieval systems. All experiments were implemented using the same \texttt{LLaMA-3.1-8B-Instruct} model with Neo4j integration and zero-shot CoT settings, ensuring fair comparison across methods.
Performance was measured using balanced accuracy (giving equal weight to true and false segments) and F1 score for detecting factual errors (Table~\ref{tab:fact_results}). Our results demonstrate that HippoRAG2 with our KG consistently outperforms baselines on Wikipedia (56.43\% accuracy, 30.48\% F1) and Common Crawl corpora, while achieving competitive results on Pes2o-Abstract. The superior performance on Wikipedia likely stems from FELM samples being partially sourced from Wikipedia content. Detailed implementation specifics and extended results are available in Appendix~\ref{sec:appendix_felm}.

\subsection{General Domain Knowledge Capabilities}

To assess AutoSchemaKG's ability to construct knowledge graphs across various domains, we evaluated it on MMLU \cite{hendrycks2021measuringmassivemultitasklanguage}, a comprehensive benchmark for LLM reasoning.
Previous research on KNN-LMs \cite{khandelwalgeneralization} suggests that retrieval-augmented generation can sometimes hinder LLMs' reasoning capabilities \cite{wang-etal-2023-knn, geng-etal-2025-great}. While we do not expect RAG to universally improve LLM performance, our findings demonstrate significant improvements in knowledge-intensive domains, even those covered in LLM training data.
Using the same retrieval and generation settings as our FELM experiments, we classified MMLU tasks into subject categories (detailed mapping in Appendix~\ref{sec:appendix_mmlu}) and focused on knowledge-intensive domains including History, Law, Religion, Philosophy/Ethics, Medicine/Health, Global Facts, Social Sciences, and Logic.

As shown in Table \ref{tab:general_benchmarks}, our ATLAS knowledge graphs enhanced performance across these domains on all tested corpora. Notably, each ATLAS variant demonstrated distinct strengths: {ATLAS-Pes2o excelled in Religion, Medicine/Health, Global Facts, and Social Sciences, reflecting its academic paper-sourced knowledge; ATLAS-Wiki showed advantages in general knowledge areas like Religion, Philosophy/Ethics, and Global Facts; while ATLAS-CC performed best in Law and History, leveraging its broader web-sourced content.
All ATLAS variants consistently outperformed both the no-retrieval baseline and Freebase-ToG in these humanities and social science domains. For example, in Law, our approach achieved a 4-point improvement over the baseline, while some other retrieval methods actually decreased performance, as shown in Table \ref{tab:general_benchmarks}. The retrieval method also matters for some specific tasks. For example, in Logic, ToG on our ATLAS knowledge graphs performs much better than all the other methods.}

The domain-specific performance pattern aligns with intuitive expectations: knowledge graphs excel in retrieving factual relationships critical for humanities and social sciences, while showing limited benefits in mathematical and technical domains where node-relation structures are less effective for capturing procedural knowledge. The complete subject analysis, including technical domains, is available in Appendix~\ref{sec:appendix_mmlu}.

\section{Further Related Work}

Knowledge graph (KG) construction transforms unstructured data into machine-readable formats. Traditional approaches using predefined schemas limit cross-domain scalability, while LLMs now enable autonomous construction through improved extraction and schema induction.
Recent advances include SAC-KG \cite{chen-etal-2024-sac}, which uses LLMs as domain experts to generate specialized KGs with high precision on million-node graphs; Docs2KG \cite{sun2024docs2kg} for heterogeneous document processing; and KAG \cite{liang2024kagboostingllmsprofessional}, which enhances multi-hop reasoning through KG-text mutual indexing.
{One of the earliest approaches to extracting schemas or concepts from entities, events, or textual descriptions leveraged lexico-syntactic patterns to automatically harvest novel lexico-semantic concepts from documents~\cite{hearst1998automated}, which enhanced WordNet~\cite{miller1995wordnet}. \citet{agirre2000enriching} further enhanced WordNet by extracting topically related concepts from web documents. Txt2onto~\cite{hawkins2022systematic} proposed a NLP-ML approach to create numerical representations of texts and use these features in a supervised learning classifier that predicts terms and concepts. Recently, due to the strong ability of LLMs to capture complex language patterns in different knowledge domains, LLMs4OL~\cite{babaei2023llms4ol} uses zero-shot prompting method with LLMs for ontology learning tasks.} Schema induction automatically derives KG structure without predefined ontologies. \citet{Hofer_2024} survey construction pipelines emphasizing ontology learning, while \citet{dash-etal-2021-open} address canonicalization using variational autoencoders and \citet{dognin-etal-2021-regen} employ reinforcement learning for bidirectional text-to-graph conversion.

\section{Conclusion}
AutoSchemaKG transforms knowledge graph construction by eliminating predefined schemas through LLM-based triple extraction and schema induction. Our methodology constructs the ATLAS family of knowledge graphs (900+ million nodes, 5.9 billion edges) with high-quality triple extraction (>95\% precision) and schema induction. This approach outperforms baselines on multi-hop QA tasks and enhances LLM factuality across diverse domains. Our work demonstrates that billion-scale knowledge graphs with dynamically induced schemas can be effectively constructed without expert intervention, providing valuable complements to parametric knowledge in large language models.

\section{Limitations}
Despite promising results, our work has several important limitations. The construction of billion-scale knowledge graphs requires substantial computational resources (78,400+ GPU hours), limiting accessibility for researchers with resource constraints. Our approach inherits biases and limitations from the underlying LLMs used for triple extraction and schema induction, potentially affecting performance in specialized domains where these models lack expertise. While achieving high semantic alignment with human-crafted schemas, our induction method still struggles with extremely technical domains requiring expert-level conceptual organization. Despite extracting billions of facts, our knowledge graphs may contain inconsistencies, contradictions, or information gaps in sparse knowledge regions. 

\section{Ethical Statement}
Our research on AutoSchemaKG and ATLAS knowledge graphs adheres to rigorous ethical standards. We exclusively utilized publicly available corpora (Dolma 1.7) with proper attribution and transparently disclosed our computational requirements (approximately 78,400 GPU hours). We acknowledge potential inherited biases from source texts and the large language models used in our pipeline, while emphasizing that our approach minimizes manual intervention that might introduce additional biases. Privacy considerations were addressed by processing only public data without extracting personally identifiable information. We candidly discuss limitations including resource constraints, LLM biases, and challenges with specialized domains. Through detailed methodological descriptions, we promote reproducibility and scientific validation. Our work aims to democratize knowledge graph construction without requiring specialized expertise or predefined schemas, potentially enabling more accurate and explainable AI systems across diverse applications and domains.

% \clearpage


\bibliography{custom}

\newpage
\clearpage

\appendix
\section{Prompts for Triple Extractions}
\label{sec:appendix}

% \begin{figure}[t]
%     \centering
%     \includegraphics[width=\linewidth]{figures/entity_prompt.pdf}
%     \caption{The figure demonstrates the prompts we use to generate the text triples describing relations between entities.}
%     \label{fig:entity_prompt}
% \end{figure}

\begin{figure}[ht]
    \small
   \begin{AIbox}{{Entity Relationship Extraction}}
Given a passage, summarize all the important entities and the relations between them in a concise manner. Relations should briefly capture the connections between entities, without repeating information from the head and tail entities. The entities should be as specific as possible. Exclude pronouns from being considered as entities. The output should strictly adhere to the following JSON format: 

[

  \quad \{
  
    \quad \quad "Head": "\{a noun\}", 
    
    \quad \quad "Relation": "\{a verb\}", 
    
    \quad \quad "Tail": "\{a noun\}", 
    
  \quad \}
  
  \quad ... 
  
]

Here is the passage:
\end{AIbox}
    \caption{The figure demonstrates the prompts we use to generate the text triples describing relations between entities.}
    \label{fig:entity_prompt}
\end{figure}

% \begin{figure}[t]
%     \centering
%     \includegraphics[width=\linewidth]{figures/event_entity_prompt.pdf}
%     \caption{The figure demonstrates the prompts we use to generate the text triples describing relations between entities and events.}
%     \label{fig:event_entity_prompt}
% \end{figure}

\begin{figure}[t]
    \centering

    \small
   \begin{AIbox}{{Event and Entity Triple Extraction}}
Please analyze and summarize the participation relations between the events and entities in the given paragraph. Each event is a single independent sentence. Additionally, identify all the entities that participated in the events. Do not use ellipses. Please strictly output in the following JSON format:

[

  \quad \{
  
    \quad \quad "Event": "\{a simple sentence describing an event\}",
    
    \quad \quad "Entity": ["\{entity 1\}", "\{entity 2\}", "..."]
    
  \quad \}
  
  \quad ... 
  
]
    \end{AIbox}

    \caption{The figure demonstrates the prompts we use to generate the text triples describing relations between entities and events.}
    \label{fig:event_entity_prompt}
\end{figure}

\begin{figure}[t]
    \centering

    \small
   \begin{AIbox}{{Event Relationship Extraction}}
Please analyze and summarize the relationships between the events in the paragraph. Each event is a single independent sentence. Identify temporal and causal relationships between the events using the following types: before, after, at the same time, because, and as a result. Each extracted triple should be specific, meaningful, and able to stand alone. Do not use ellipses. The output should strictly adhere to the following JSON format:

[

  \quad \{
  
    \quad \quad "Head": "\{a simple sentence describing the event 1\}",
    
    \quad \quad "Relation": "\{temporal or causality relation between the events\}",
    
    \quad \quad "Tail": "\{a simple sentence describing the event 2\}"
    
  \quad \}
  
  \quad ... 
  
]
    \end{AIbox}

    \caption{The figure demonstrates the prompts we use to generate the text triples describing relations between events.}
    \label{fig:event_prompt}
\end{figure}

% \begin{figure}[t]
%     \centering
%     \includegraphics[width=\linewidth]{figures/event_prompt.pdf}
%     \caption{The figure demonstrates the prompts we use to generate the text triples describing relations between events.}
%     \label{fig:event_prompt}
% \end{figure}

The prompts used for extracting entity-entity, entity-event, and event-event triples are given in Figure \ref{fig:entity_prompt}, Figure \ref{fig:event_entity_prompt}, and Figure \ref{fig:event_prompt} respectively.

\section{Implementation Details of Knowledge Graph Construction Framework}

In this section, we elaborate on the process of fully automating knowledge graph construction. Given a collection of $n$ documents $D = \{D_1, D_2, D_3, \dots, D_n\}$, our method systematically builds the graph.

\subsection{Triple Extraction}
\label{sec:triple_extraction}

Our approach to triple extraction employs a multi-phase pipeline that utilizes the generative power of Large Language Models (LLMs) to convert unstructured text into structured knowledge triples, drawing from the Dolma corpus \cite{DBLP:journals/corr/abs-2402-00159}. This pipeline systematically extracts three categories of relationships—Entity-Entity, Entity-Event, and Event-Event—forming the foundation of a comprehensive knowledge graph. Designed for scalability and resilience, the method incorporates batch processing, text segmentation, and robust output parsing to efficiently process large-scale datasets.

The extraction unfolds across three sequential stages, each tailored to a specific relationship type, leveraging a single LLM guided by distinct prompts to produce structured outputs in JSON format. To manage the constraints of LLM input capacity, denoted as $L_{max}$ tokens, we preprocess the text corpus to ensure compatibility, segmenting documents as needed and organizing them into batches for efficient processing. This section outlines the preprocessing strategy, the staged extraction process, and key implementation details.

\subsubsection{Text Preprocessing and Data Organization}
Given a corpus $D = \{D_1, D_2, \dots, D_n\}$ of $n$ documents, we first filter the dataset to include only English-language texts, identified through metadata or assumed if unspecified, to match the linguistic capabilities of our LLMs. To adhere to the token limit $L_{max}$, we account for an instructional prompt length, $L_{inst}$ derived from empirical observations. The maximum token length per text segment, $C_{max}$, is calculated as: \(C_{max} =  (L_{max} - L_{inst}) \)

Documents exceeding $C_{max}$ are divided into smaller chunks, each tagged with a unique identifier and metadata to maintain traceability. This segmentation ensures that inputs remain within the LLM’s token capacity, preserving contextual integrity without truncation.

The preprocessed text is then grouped into batches of size $B$, utilizing a custom data management framework that integrates with standard dataset loading tools. Tokenization is applied to each batch, adjusting for padding and truncation to produce consistent input representations suitable for LLM processing.

\subsubsection{Stage 1: Extraction of Entity-Entity Relationships}
In the initial stage, as shown in Figure \ref{fig:entity_prompt}, we extract Entity-Entity relationships, identifying connections between named entities such as individuals, organizations, or locations. For each batch, we prepend a system prompt, $P_{EE}$, which instructs the LLM to detect entities and their interrelations, followed by the segmented text. The LLM generates a response in JSON format, which is subsequently decoded and parsed. The parsing process isolates the structured content by locating the model’s answer start token, $T_{start}$, repairs any malformed JSON, and extracts a list of dictionaries. Each dictionary represents a triple $(e_1, r, e_2)$, where $e_1, e_2 \in V_N$ are entity nodes and $r \in R$ is a relation type. If parsing encounters errors, an empty list is returned to ensure pipeline continuity .

\subsubsection{Stage 2: Extraction of Entity-Event Relationships}
The second stage focuses on Entity-Event relationships, as shown in Figure \ref{fig:event_entity_prompt}, linking entities to specific occurrences or events. Using the original text segments from Stage 1, we apply a new prompt, $P_{EV}$, directing the LLM to identify events and their associated entities. The generation and parsing steps mirror those of Stage 1, producing triples of the form $(e, r, v)$ or $(v, r, e)$, where $e \in V_N$, $v \in V_E$ (event nodes), and $r \in R$. This bidirectional extraction captures both entities involved in events and events tied to entities, enhancing the graph’s relational depth.

\subsubsection{Stage 3: Extraction of Event-Event Relationships}
The third stage targets Event-Event relationships, as shown in Figure \ref{fig:event_prompt},, detecting causal, temporal, or logical connections between events. A specialized prompt, $P_{VV}$, is applied to the text segments, prompting the LLM to generate triples of the form $(v_1, r, v_2)$, where $v_1, v_2 \in V_E$ and $r \in R$. The parsing process follows the same methodology, repairing JSON outputs as needed. To accommodate potentially intricate event descriptions, we extend the generation limit to $L_{ext} = \alpha \cdot L_{max}$, where $\alpha > 1$ is a scaling factor, ensuring comprehensive capture of event interactions.

\subsubsection{Implementation Considerations}
The pipeline supports a variety of LLMs, including models from Google, Meta, Mistral, Microsoft, and others, configured with optimized precision settings (e.g., bfloat16 or float16) and enhanced with acceleration techniques where applicable. Deployment occurs on GPUs, with input-output formatting governed by model-specific chat templates, $T_{chat}$, to ensure compatibility. The extracted triples, along with their corresponding texts and metadata, are serialized into JSON files per batch, enabling subsequent schema induction and evaluation.

This multi-stage pipeline achieves thorough triple extraction by addressing each relationship type systematically, harnessing the LLM’s generative capabilities within a scalable and fault-tolerant framework. The use of variables such as $L_{max}$, $C_{max}$, and $B$ ensures flexibility across different models and datasets, reinforcing the methodology’s adaptability and generalizability.

\subsection{Schema Induction}
\label{sec:schema_induction}
Following the extraction of knowledge triples, our methodology advances to schema induction, a critical step that abstracts specific entities, events, and relations into generalized types to form a coherent and adaptable schema for the knowledge graph. This process leverages the contextual understanding of Large Language Models (LLMs) to generate conceptual phrases that represent the types or related concepts of each graph element, enabling the graph to scale across diverse domains without manual schema design. The induced schema aligns with the formal definition of a knowledge graph $G = (V, E, C, \phi, \psi)$, where $C$ denotes the set of concepts, and $\phi$ and $\psi$ map nodes and relations to subsets of $C$, respectively.

Our schema induction pipeline processes the triples extracted from the Dolma corpus \cite{DBLP:journals/corr/abs-2402-00159}, organizing them into batches and employing a generative approach to derive abstract representations. The process targets three components—events ($V_E$), entities ($V_N$), and relations ($R$)—producing a set of conceptual phrases for each, which collectively form the concept set $C$. This section outlines the abstraction methodology, the role of context in entity conceptualization, and key implementation details.

\begin{figure}[ht]
    \small
    \begin{AIbox}{{Abstract Event Phrase Generation}}
I will give you an EVENT. You need to give several phrases containing 1-2 words for the ABSTRACT EVENT of this EVENT.
You must return your answer in the following format: phrases1, phrases2, phrases3,...
You can't return anything other than answers.
These abstract event words should fulfill the following requirements:

\quad 1. The ABSTRACT EVENT phrases can well represent the EVENT, and it could be the type of the EVENT or the related concepts of the EVENT. 

\quad 2. Strictly follow the provided format, do not add extra characters or words.

\quad 3. Write at least 3 or more phrases at different abstract level if possible.

\quad 4. Do not repeat the same word and the input in the answer.

\quad 5. Stop immediately if you can't think of any more phrases, and no explanation is needed.

Examples:

EVENT: A man retreats to mountains and forests
Your answer: retreat, relaxation, escape, nature, solitude
            
EVENT: A cat chased a prey into its shelter
Your answer: hunting, escape, predation, hidding, stalking

EVENT: Sam playing with his dog
Your answer: relaxing event, petting, playing, bonding, friendship

EVENT: [EVENT]
Your answer:
    \end{AIbox}
    \caption{This figure shows the prompt used for generating the concepts for an event. 
    }
    \label{fig:event_conceptualization_prompt}
\end{figure}

\subsubsection{Abstraction Methodology}
The schema induction begins by categorizing the nodes and edges of the knowledge graph $G$ into events, entities, and relations. For each category, we process the elements in batches of size $B_s$ to optimize computational efficiency and scalability. The LLM is prompted with tailored instructions to generate a list of phrases, each containing one to two words, that abstractly represent the input element. These phrases must satisfy several criteria: they should encapsulate the element’s type or related concepts, vary in abstraction level, and avoid repetition or inclusion of the original input term. For each element, a minimum of three phrases is targeted, though more may be generated depending on the LLM’s output.

For events ($v \in V_E$), the prompt directs the LLM to identify abstract event types or related notions. For example, an event such as "Sam playing with his dog" might yield phrases like "playing," "bonding," and "relaxing event," reflecting different levels of generality. For entities ($e \in V_N$), the prompt similarly elicits abstract entity types, augmented by contextual information derived from the graph structure, as detailed below. Relations ($r \in R$) are abstracted into phrases that capture their semantic essence, such as transforming "participated in" into "engage in," "attend," and "involve in." The LLM generates these outputs in a structured format, which we parse into lists of phrases, forming the mappings $\phi(v)$, $\phi(e)$, and $\psi(r)$ to the concept set $C$.

The abstraction process operates in batches, processing $B_s$ elements simultaneously. The input prompts are tokenized to a maximum length $L_{tok}$, and the LLM generates responses under controlled parameters (e.g., temperature $\tau$ and top-$p$ sampling with probability $p$) to balance creativity and coherence. The resulting phrases are stored alongside their corresponding elements, ensuring traceability and enabling subsequent analysis.

\subsubsection{Contextual Enhancement for Entities}
As shown in Figure \ref{fig:entity_conceptualization_prompt}, to enhance the accuracy of entity abstraction, we incorporate contextual information extracted from the knowledge graph. For each entity $e \in V_N$, we examine its neighboring nodes—predecessors and successors—along with their associated relations. A subset of these neighbors, limited to $N_{ctx}$ (e.g., one predecessor and one successor), is randomly sampled to construct a context string. This string concatenates the neighbor’s identity and relation (e.g., "neighbor1 relation1, relation2 neighbor2"), providing the LLM with additional semantic cues. For instance, an entity "Black Mountain College" with context "started by John Andrew Rice" might yield phrases like "college," "school," and "liberal arts college." This contextual enrichment ensures that the abstracted types are grounded in the entity’s role within the graph, improving the schema’s relevance and specificity.

\begin{figure}[ht]
    \small
    \begin{AIbox}{{Abstract Entity Phrase Generation}}
I will give you an ENTITY. You need to give several phrases containing 1-2 words for the ABSTRACT ENTITY of this ENTITY.
You must return your answer in the following format: phrases1, phrases2, phrases3,...
You can't return anything other than answers.
These abstract intention words should fulfill the following requirements:

\quad 1. The ABSTRACT ENTITY phrases can well represent the ENTITY, and it could be the type of the ENTITY or the related concepts of the ENTITY.

\quad 2. Strictly follow the provided format, do not add extra characters or words.

\quad 3. Write at least 3 or more phrases at different abstract level if possible.

\quad 4. Do not repeat the same word and the input in the answer.

\quad 5. Stop immediately if you can't think of any more phrases, and no explanation is needed.

Examples:

ENTITY: Soul
CONTEXT: premiered BFI London Film Festival, became highest-grossing Pixar release
Your answer: movie, film

ENTITY: Thinkpad X60
CONTEXT: Richard Stallman announced he is using Trisquel on a Thinkpad X60
Your answer: Thinkpad, laptop, machine, device, hardware, computer, brand

ENTITY: Harry Callahan
CONTEXT: bluffs another robber, tortures Scorpio
Your answer: person, Amarican, character, police officer, detective

ENTITY: Black Mountain College
CONTEXT: was started by John Andrew Rice, attracted faculty
Your answer: college, university, school, liberal arts college

ENTITY: 1st April
CONTEXT: Utkal Dibas celebrates
Your answer: date, day, time, festival

ENTITY: [ENTITY]
CONTEXT: [CONTEXT]
Your answer:
    \end{AIbox}
    \caption{This figure shows the conceptualization prompts for entities enhanced with context.}
    \label{fig:entity_conceptualization_prompt}
\end{figure}

\begin{figure}[ht]
    \small
    \begin{AIbox}{{Abstract Relation Phrase Generation}}
I will give you a RELATION. You need to give several phrases containing 1-2 words for the ABSTRACT RELATION of this RELATION.
You must return your answer in the following format: phrases1, phrases2, phrases3,...
You can't return anything other than answers.
These abstract intention words should fulfill the following requirements:

\quad 1. The ABSTRACT RELATION phrases can well represent the RELATION, and it could be the type of the RELATION or the simplest concepts of the RELATION.

\quad 2. Strictly follow the provided format, do not add extra characters or words.

\quad 3. Write at least 3 or more phrases at different abstract level if possible.

\quad 4. Do not repeat the same word and the input in the answer.

\quad 5. Stop immediately if you can't think of any more phrases, and no explanation is needed.

Examples:

RELATION: participated in
Your answer: become part of, attend, take part in, engage in, involve in

RELATION: be included in
Your answer: join, be a part of, be a member of, be a component of

RELATION: [RELATION]
Your answer:
    \end{AIbox}
    \caption{This figure shows the conceptualization prompts for relations enhanced with context.}
    \label{fig:relation_conceptualization_prompt}
\end{figure}

Events and relations, as shown in Figure \ref{fig:event_conceptualization_prompt} and Figure \ref{fig:relation_conceptualization_prompt}, by contrast, rely solely on their textual descriptions without additional context, as their abstraction focuses on inherent semantics rather than graph connectivity. This distinction reflects the differing roles of nodes and edges in the knowledge graph structure.

\subsubsection{Implementation Details}
The schema induction pipeline processes a graph $G$ serialized from the triple extraction phase, typically stored in a binary format and loaded into memory. The elements are partitioned into batches, with the option to apply slicing (dividing the workload into $S_{total}$ slices and processing the $S_{slice}$-th portion) for distributed computation. If a sample size $N_{sample}$ is specified, a random subset of batches is selected to reduce processing time during experimentation.

The LLM, configured with a precision setting (e.g., float16) and optimized with acceleration techniques, operates on a GPU to handle the batched inference efficiently. Prompts are formatted using a model-specific chat template, $T_{chat}$, ensuring compatibility with the LLM’s input-output conventions. The generated phrases are written to a CSV file, with each row recording the original element, its abstracted phrases, and its type (event, entity, or relation). Post-processing aggregates these phrases to compute the unique concepts in $C$, providing statistics on the schema’s coverage, such as the number of distinct event types, entity types, and relation types.

This approach yields a flexible and automated schema, mapping each node $v \in V$ and relation $r \in R$ to a subset of concepts in $C$ via $\phi$ and $\psi$. By abstracting specific instances into general types, the induced schema enhances the knowledge graph’s adaptability, supporting downstream applications across varied domains without requiring manual curation.

\begin{figure}[ht]
    \small
    \begin{AIbox}{{Multiple-Choice Question Generation and Answering}}
\textbf{MCQ Generation Prompt:}

You are an expert in generating multiple-choice questions (MCQs) from scientific texts.
Your task is to generate 5 multiple-choice questions based on the following passage.

Each question should:

\quad - Focus on factual claims, numerical data, definitions, or relational knowledge from the passage.

\quad - Have 4 options (one correct answer and three plausible distractors).

\quad - Clearly indicate the correct answer.

The output should be in JSON format, with each question as a dictionary containing:

\quad - "question": The MCQ question.

\quad - "options": A list of 4 options (e.g., ["A: ..", "B: ..", "C: ..", "D: .."]).

\quad - "answer": The correct answer (e.g., "A").

Output Example:

[

  \quad \{
  
    \quad \quad "question": "What is the primary role of a catalyst in a chemical reaction?",
    
    \quad \quad "options": [
    
      \quad \quad \quad "A: To make a thermodynamically unfavorable reaction proceed",
      
      \quad \quad \quad "B: To provide a lower energy pathway between reactants and products",
      
      \quad \quad \quad "C: To decrease the rate of a chemical reaction",
      
      \quad \quad \quad "D: To change the overall reaction itself"
      
    \quad \quad ],
    
    \quad \quad "answer": "B"
    
  \quad \}
  
]

Passage:
\{passage\}

\hrulefill

\textbf{MCQ Answering Prompt:}

Given the contexts or evidences:
\{contexts\}

Here is a multiple-choice question:

Question: \{question\}

Options:
A. \{options\_0\}
B. \{options\_1\}
C. \{options\_2\}
D. \{options\_3\}

Please select the correct answer by choosing A, B, C, or D. Respond with only the letter of your choice.
    \end{AIbox}
    \caption{The prompts for generating and answering MCQ questions for evaluating knowledge retention in knowledge graph.}
    \label{fig:mcq_prompts}
\end{figure}

\section{Experiment Settings of Schema Accuracy}\label{appendix:sa_settings}
\subsection{Datasets}\label{appendix:sa_dataset}

\textbf{Entity Typing.} We conduct experiments on the typed entities of two real-world knowledge graphs, FB15kET~\cite{bordes2013translating} and YAGO43kET~\cite{moon2017learning}, which are the subsets of Freebase~\cite{bollacker2008freebase} and YAGO~\cite{suchanek2007yago}, respectively. The types of entities are collected from \cite{10.1145/3132847.3133095}. There are 3,584 and 45,182 entity types in FB15kET and YAGO43kET, respectively. We utilize the entities in the testing sets of these two datasets with their types as ground truths to validate the entity induction performance of our schema induction method.

\textbf{Event Typing.} We conduct experiments on the typed events of wikiHow~\cite{koupaee2018wikihow}, which is an online community contains a collection of professionally edited how-to guideline articles. The types of events are collected by P2GT~\cite{chen2020you}. There are 625 event types among 12,795 events. We utilize the events in the testing set of wikiHow with their types as ground truths to validate the event induction performance of our schema induction method.

\textbf{Relation Typing.} There are no datasets designed for the relation typing task, so here we make use of the domain segments separated by "/" in FB15kET~\cite{bordes2013translating} to extract the types. These domain segments serve as ground truth types, with the last domain component functioning as the relation itself. There are 607 relation types among 1,345 relations in FB15kET. We utilize the relations in the testing set of FB15kET with their types as ground truths to validate the relation induction performance of our schema induction method.

\subsection{Metrics}\label{appendix:sa_metric}
We employ \textbf{BertScore-Recall} and \textbf{BertScore-Coverage} as the evaluation metrics, which are denoted as \textbf{BS-R} and \textbf{BS-C} respectively. They are used to calculate how many types in each instance or entire testing set are recalled by our schema induction method. The BertScore~\cite{zhang2019bertscore}, which is denoted as \textbf{BS}, between each pair of type and induced schema are calculated as follows:
\begin{equation}
\text{BertRecall} = \frac{1}{|t|} \sum_{t_i \in t}\max_{\hat{t}_j \in \hat{t}} \mathbf{x}_{t_i}^{\top}\mathbf{x}_{\hat{t}_j},
    \label{eq:16}
\end{equation}
\begin{equation}
\text{BertPrec} = \frac{1}{|\hat{t}|} \sum_{\hat{t}_i \in \hat{t}} \max_{t_j \in t} \mathbf{x}_{t_j}^{\top}\mathbf{x}_{\hat{t}_i},
    \label{eq:17}
\end{equation}
\begin{equation}
\text{BS}(t, \hat{t}) = 2\frac{\text{BertRecall} \cdot \text{BertPrec}}{\text{BertRecall} + \text{BertPrec}},
    \label{eq:16}
\end{equation}
where $t$ and $\hat{t}$ represents the tokens of a ground truth type and induced schema, respectively. The embedding vector of each token $t_i$ or $\hat{t}_j$ of a type $t$ or induced schema $\hat{t}$ is denoted as $\mathbf{x}_{t_i}$ and $\mathbf{x}_{\hat{t}_j}$, which are obtained with RoBERTa~\cite{liu2019roberta}. Then the BS-R and BS-C can be calculated as follows:
\begin{equation}
\text{BS-R}(\mathcal{T}, \hat{\mathcal{T}}) = \frac{1}{|\hat{\mathcal{T}}|} \sum_{\hat{t} \in \hat{\mathcal{T}}}\max_{t \in \mathcal{T}} \text{BS}(t,\hat{t}),
    \label{eq:17}
\end{equation}
\begin{equation}
\text{BS-C}(\mathcal{S}_{t}, \mathcal{S}_{\hat{t}}) = \frac{1}{|\mathcal{S}_{\hat{t}}|} \sum_{\hat{t} \in \mathcal{S}_{\hat{t}}}\max_{t \in \mathcal{S}_{t}} \text{BS}(t,\hat{t}),
    \label{eq:18}
\end{equation}
where $\mathcal{T}$ and $\hat{\mathcal{T}}$ represent a set of ground truth types and induced schemas in each testing instance, respectively. Similarly, $\mathcal{S}_{t}$ and $\mathcal{S}_{\hat{t}}$ denote the set of ground truth types and induced schemas across the entire testing set, respectively.

\section{Case Study Examples}

\begin{figure*}[h]
    \centering
    \includegraphics[width=\linewidth]{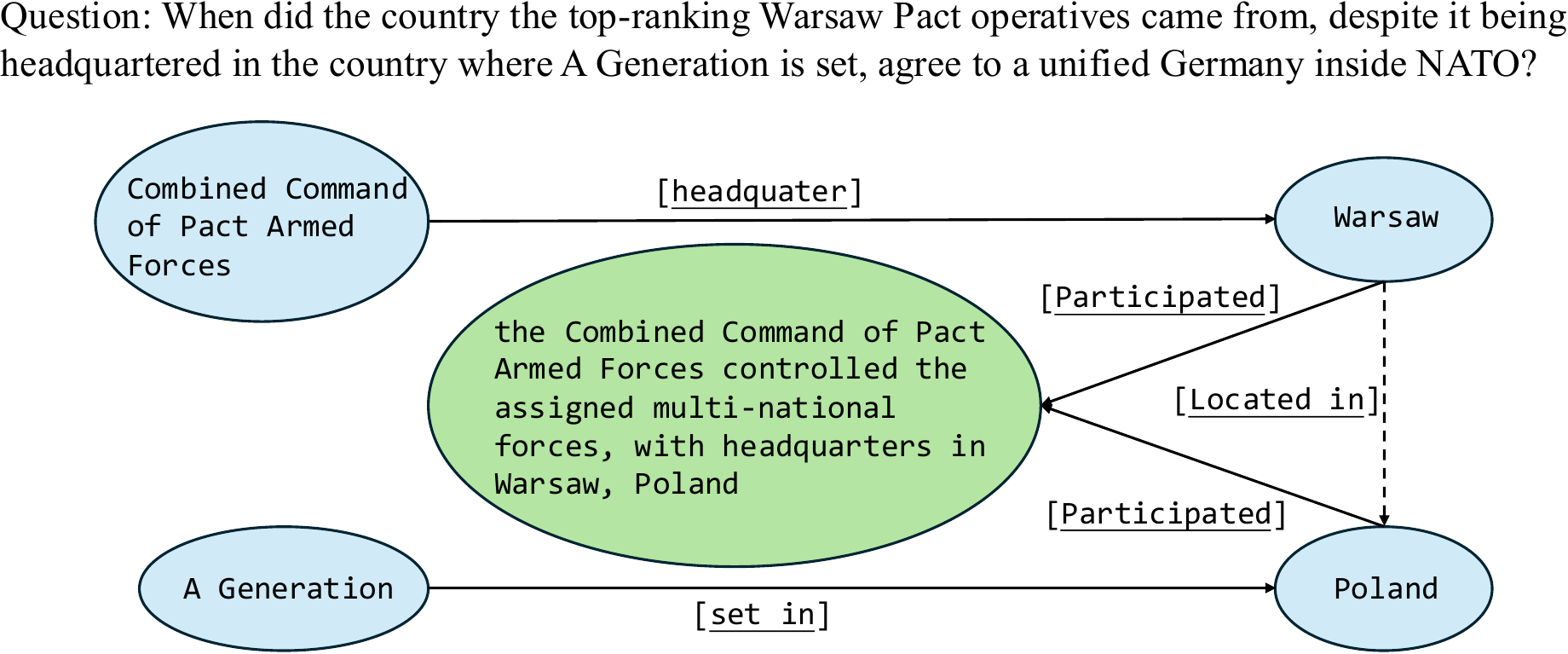}
    \caption{ Event Node (green) offers enriched context over triplets (blue); dotted line indicates missing edge}
    \label{fig:event-example}
\end{figure*}

\begin{figure*}[h]
    \centering
    \includegraphics[width=\linewidth]{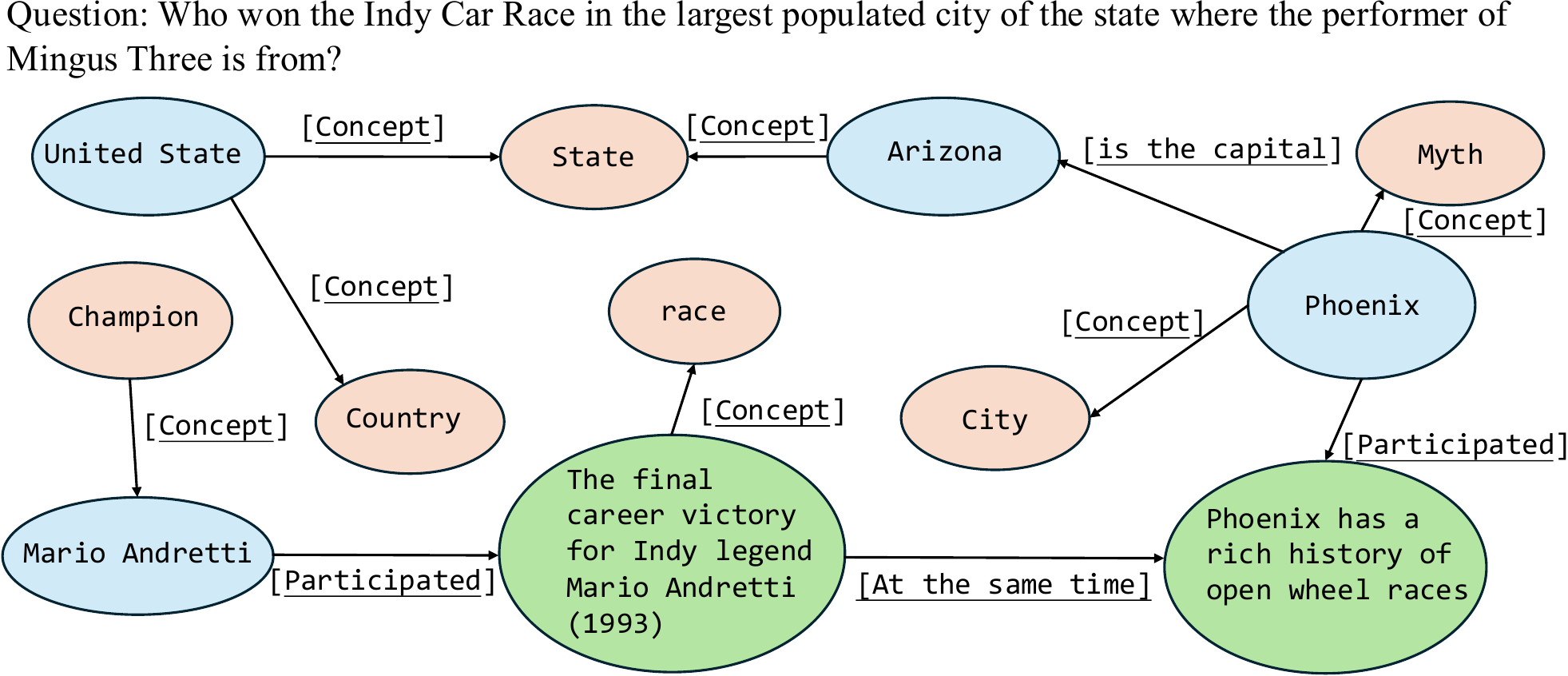}
    \caption{ Concept nodes (orange) provide alternate pathways to access information beyond entities and events.}
    \label{fig:concept-example}
\end{figure*}

Figures \ref{fig:event-example} and \ref{fig:concept-example} demonstrate specific cases where events and concepts are crucial for effective knowledge graph utilization in retrieval-augmented generation. Figure \ref{fig:event-example} illustrates how event nodes provide essential contextual information that entity-only representations miss, while Figure \ref{fig:concept-example} showcases how concept nodes establish semantic bridges across otherwise disconnected subgraphs, enabling more comprehensive reasoning for complex questions.

\section{Algorithm for RAG}

We include all the algorithms used in our RAG evaluation on various graphs constructed by AutoSchemaKG. Algorithm \ref{alg:tog} presents the Think-on-Graph reasoning method that leverages our knowledge graphs for multi-hop question answering. For adapting our entity-event-concept graphs at different scales, we implemented two variants of HippoRAG2: Algorithm \ref{alg:hipporag_small} for smaller, more focused graph traversal, and Algorithm \ref{alg:hipporag_large} for large-scale graph exploration with optimized memory management. These adaptations enable efficient navigation of the rich semantic structures in our ATLAS knowledge graphs.

\begin{algorithm*}
\caption{Think on Graph (ToG) \cite{sun2024thinkongraph} for Question Answering}
\label{alg:tog}
\begin{algorithmic}[1]
\Require Knowledge Graph $G$, Query $q$, Top-$N$ parameter, Maximum depth $D_{max}$
\Ensure Answer to query $q$
\State Extract entities from query $q$ using NER
\State Retrieve top-$k$ initial nodes from $G$ based on entity similarity
\State Let $P \gets$ set of paths, each containing a single initial node
\State $D \gets 0$ \Comment{Current search depth}
\While{$D \leq D_{max}$}
    \State $P \gets \texttt{Search}(q, P, G)$ \Comment{Expand paths by one hop}
    \State $P \gets \texttt{Prune}(q, P, N)$ \Comment{Keep top-$N$ most relevant paths}
    \If{$\texttt{Reasoning}(q, P)$ determines paths sufficient}
        \State \Return $\texttt{Generate}(q, P)$ \Comment{Generate answer using paths}
    \EndIf
    \State $D \gets D + 1$
\EndWhile
\State \Return $\texttt{Generate}(q, P)$ \Comment{Generate answer using best available paths}

\Procedure{Search}{$q, P, G$}
    \State $P_{new} \gets \emptyset$
    \For{each path $p \in P$}
        \State $e_{tail} \gets$ last entity in path $p$
        \State $S \gets$ successors of $e_{tail}$ in $G$ not already in $p$
        \State $R \gets$ predecessors of $e_{tail}$ in $G$ not already in $p$
        \If{$S = \emptyset$ and $R = \emptyset$}
            \State $P_{new} \gets P_{new} \cup \{p\}$ \Comment{Keep dead-end paths}
        \Else
            \For{each node $n \in S$}
                \State $r \gets$ relation from $e_{tail}$ to $n$ in $G$
                \State $P_{new} \gets P_{new} \cup \{p + [r, n]\}$ \Comment{Extend path forward}
            \EndFor
            \For{each node $n \in R$}
                \State $r \gets$ relation from $n$ to $e_{tail}$ in $G$
                \State $P_{new} \gets P_{new} \cup \{p + [r, n]\}$ \Comment{Extend path backward}
            \EndFor
        \EndIf
    \EndFor
    \State \Return $P_{new}$
\EndProcedure

\Procedure{Prune}{$q, P, N$}
    \State Score each path in $P$ using LLM relevance assessment (1-5 scale)
    \State Sort paths by decreasing score
    \State \Return top-$N$ highest scoring paths
\EndProcedure

\Procedure{Reasoning}{$q, P$}
    \State Extract triples from paths in $P$
    \State Ask LLM if triples are sufficient to answer $q$ (Yes/No)
    \State \Return True if answer is "Yes", False otherwise
\EndProcedure

\Procedure{Generate}{$q, P$}
    \State Extract triples from paths in $P$
    \State Prompt LLM with triples and query $q$ to generate answer
    \State \Return generated answer
\EndProcedure
\end{algorithmic}
\end{algorithm*}

\begin{algorithm*}
\caption{HippoRAG2 \cite{gutierrez2025rag}\\ General algorithm follows the original implementation, while we modify the initialization of graph and embeddings.}
\label{alg:hipporag_small}
\begin{algorithmic}[1]

\Function{init}{graph\_type}
    \If{graph\_type is \textbf{entity}}
        \State $\text{graph, embedding} \gets  Graph(entity), Embeddings(entity)$
    \ElsIf{graph\_type is \textbf{entity+event}}
        \State $\text{graph, embedding} \gets  Graph(entity,event), Embeddings(entity,event)$
    \ElsIf{graph\_type is \textbf{entity+event+concept}}
        \State $\text{graph, embedding} \gets  Graph(entity,event,concept), Embeddings(entity,event,concept)$
    \EndIf
\EndFunction

\Function{query2edge}{query, topN}
    \State $Q_{emb} \gets$ Retriever(query)
    \State $S = Q\cdot W_e$ \Comment{Calculate similarity scores with precomputed edge embeddings}
    \State $E = \text{argsort}_i(S)[:N]$\Comment{Select topN edges based on scores}
    \State $\text{filtered\_edges} \gets \text{LLM\_filter}(E)$ \Comment{Filter edges using Large Language Model}
    \State $\text{mapped\_edges} \gets \text{Map\_edges}(\text{filtered\_edges})$ \Comment{Map filtered edges to original edges}
    \State $\text{return\_node\_scores} \gets \text{Calculate\_node\_scores}(\text{mapped\_edges})$
    \State \Return \text{return\_node\_scores}
\EndFunction

\Function{query2passage}{query, weight\_adjust}
    \State $Q_{pass} \gets \text{Encode}(query)$ \Comment{Encode query into passage representation}
    \State $S_{text} \gets \text{Similarity\_Scores}(Q_{pass}, \text{text\_embeddings})$
    \State \Return \text{Scores\_Dictionary}($S_{text}$)
\EndFunction

\Function{retrieve\_personalization\_dict}{query, topN}
     \State $\text{node\_dict} \gets \text{query2edge}(query, topN)$
    \State $\text{text\_dict} \gets \text{query2passage}(query, \text{weight\_adjust})$
    \State \Return \text{node\_dict, text\_dict}
\EndFunction

\Function{retrieve\_passages}{query, topN}
     \State $\text{node\_dict, text\_dict} \gets \text{retrieve\_personalization\_dict}(query, topN)$
    \If{\text{node\_dict is empty}}
        \State \Return \text{TopN\_Text\_Passages}(text\_dict)
    \Else
        \State \text{personalization\_dict} $\gets$ \{\text{node\_dict, text\_dict}\}
        \State $\text{page\_rank\_scores} \gets \text{PageRank}(\text{personalization\_dict})$
        \State \Return \text{TopN\_Passages}(page\_rank\_scores)
    \EndIf
\EndFunction
\end{algorithmic}
\label{fig:hipporag2_algo}
\end{algorithm*}

\begin{algorithm*}
\caption{LargeKGRetriever\\ A variant of HippoRAG2 \cite{gutierrez2025rag}, optimized with dynamic graph sampling and common word filtering}
\label{alg:hipporag_large}
\begin{algorithmic}[1]
\Function{init}{graph\_type}
    \State \text{keyword} $\gets$ \text{Default\_graph\_keyword} \Comment{Keyword can be cc, pes2o, wiki}
    \State \text{Initialize\_resources(keyword)}
    \State \text{Load\_node\_and\_edge\_indexes()}
\EndFunction

\Function{retrieve\_topk\_nodes}{query, top\_k\_nodes}
    \State $\text{entities} \gets \text{LLM\_NER(query)}$
    \State $\text{KG\_entities} \gets \text{Encode\_and\_Search(entities, \text{FAISS\_index})}$
    \State $\text{filtered\_keywords} \gets \text{LLM\_filter}(\text{KG\_entities})$
    \State \Return \text{filtered\_keywords}
\EndFunction

\Function{retrieve\_personalization\_dict}{query, number\_of\_source\_nodes}
    \State $\text{topk\_nodes} \gets \text{retrieve\_topk\_nodes}(query, \text{number\_of\_source\_nodes})$
    \If{$\text{topk\_nodes} == \{\}$}
        \State \Return $\{\}$
    \EndIf
    \State \text{Update personalization dictionary with topk\_nodes}
    \State \Return \text{Personalization dictionary}
\EndFunction

\Function{pagerank}{personalization\_dict, topN, sampling\_area}
    \State $G_{\text{Sample}} \gets$ Random Walk with Restart Sampling
    \State $Scores$ = PageRank($G_{\text{Sample}}$, personalization\_dict)
    \State topN\_nodes = $\text{argsort}_i(Scores)[:N]$
    \For{node in topN nodes}
        \State Connected\_Passage += node.score 
    \EndFor
    \State \Return \text{TopN\_Ranked\_Passages}
\EndFunction

\Function{retrieve\_passages}{query, topN, number\_of\_source\_nodes, sampling\_area}
    \State \text{personalization\_dict} $\gets$ \text{retrieve\_personalization\_dict}(query, \text{number\_of\_source\_nodes})
    \If{$\text{personalization\_dict}$ is empty} 
        \State \Return $\{\}$, [0]
    \EndIf
    \State $\text{topN\_passages} \gets \text{pagerank}(\text{personalization\_dict}, \text{topN}, \text{sampling\_area})$
    \State \Return $\text{topN\_passages}$
\EndFunction
\end{algorithmic}
\end{algorithm*}
% \begin{tcolorbox}[colback=blue!10, colframe=blue!50!black]
%   This is a text inside a blue colored box.
% \end{tcolorbox}

% % More customized box
% \begin{tcolorbox}[
%   title=Important Note,
%   colback=red!5,
%   colframe=red!75!black,
%   colbacktitle=red!75!black,
%   fonttitle=\bfseries
% ]
%   This is an important note in a red box with a title.
% \end{tcolorbox}

% \begin{mdframed}[linecolor=green!50!black, backgroundcolor=green!10]
%   This is text in a green frame with light green background.
% \end{mdframed}

\section{The Recall Metrics in Opendomain QA Tasks}

We also use Retrieval Quality metrics at $k \in \{2, 5\}$: 
% $\text{PR@}k = \mathbf{1}[|D_k \cap S| \geq 1]$ 
$\text{PR@}k = |D_k \cap S| / |S| $
% and $\text{FR@}k = \mathbf{1}[S \subseteq D_k]$, 
where PR@$k$ (Partial Recall) measures the fraction supporting document is in top-$k$, 
% FR@$k$ (Full Recall) measures if all supporting documents are in top-$k$, 
$D_k$ is the set of top-$k$ retrieved documents, and $S$ is the set of supporting documents. For multi-hop QA datasets (HotpotQA, 2WikiMultihopQA, MuSiQue), these retrieval metrics are crucial as they measure how effectively our system retrieves the evidence needed for multi-step reasoning.

Questions in datasets like 2WikiMultihopQA \cite{ho-etal-2020-constructing} and HotpotQA \cite{yang2018hotpotqa} tend to be more entity-centric, with relationships and entities more explicitly represented, which aids retrievers in easily locating relevant subgraphs. In contrast, MuSiQue \cite{trivedi2021musique}, due to its questions' increased complexity in both description and multi-hop nature, poses greater challenges for retrieval. Additionally, differences in graph construction cause the retrievers to perform differently across datasets.
\begin{table*}[]
\small
\centering
\begin{tabular}{@{}l|cc|cc|cc@{}}
\toprule
\textbf{Model/Dataset} & \multicolumn{2}{c|}{\textbf{MuSiQue}} & \multicolumn{2}{c|}{\textbf{2Wiki}} & \multicolumn{2}{c}{\textbf{HotpotQA}} \\ \midrule
\textbf{Metric} & \textbf{Recall@2} & \textbf{Recall@5} & \textbf{Recall@2} & \textbf{Recall@5} & \textbf{Recall@2} & \textbf{Recall@5} \\ \midrule
\rowcolor[gray]{0.9}\multicolumn{7}{l}{\textit{Baseline Retrievers }} \\ \midrule
Contriever & 34.8 & 46.6 & 46.6 & 57.5 & 58.4 & 75.3 \\
BM25 & 32.4 & 43.5 & 55.3 & 65.3 & 57.3 & 74.8 \\ \midrule

\rowcolor[gray]{0.9}\multicolumn{7}{l}{\textit{LLM Embeddings}} \\ \midrule
GTE-Qwen2-7B-Instruct & 48.1 & 63.6 & 66.7 & 74.8 & 75.8 & 89.1  \\
GritLM-7B & 49.7 & 65.9 & 67.3 & 76.0 & 79.2 & 92.4  \\
NV-Embed-v2 (7B) &  52.7 & 69.7 & 67.1 & 76.5 & 84.1 & 94.5 \\ \midrule

\rowcolor[gray]{0.9}\multicolumn{7}{l}{\textit{Existing Graph-based RAG Methods}} \\ \midrule
RAPTOR  (Llama-3.3-70B-Instruct) & 47.0 & 57.8 & 58.3 & 66.2 & 76.8 & 86.9 \\
HippoRAG (Llama-3.3-70B-Instruct) & 41.2 & 53.2 & 71.9 & 90.4 & 60.4 & 77.3 \\
HippoRAG2 (Llama-3.3-70B-Instruct) & 56.1 & {74.7} & 76.2 & 90.4 & 83.5 & 96.3 \\ 
\midrule
\rowcolor[gray]{0.9} \multicolumn{7}{l}{\textit{AutoSchemaKG (\texttt{LLama-3.1-8B-Instruct}) + HippoRAG1}} \\ \midrule
Entity-KG (Llama-3-8B-Instruct) & 41.37 & 51.08 & {61.72} & {75.45} & 51.89 & 65.95 \\
Entity-Event-KG (Llama-3-8B-Instruct) & {41.28} & {51.12} & 61.37& 74.56 & {51.31} & {65.93} \\
Full-KG (Llama-3-8B-Instruct) & 40.78 & 50.36 & 61.08 & 71.9 & 52.8 & 65.4 \\ \midrule
\rowcolor[gray]{0.9} \multicolumn{7}{l}{\textit{AutoSchemaKG (\texttt{LLama-3.1-8B-Instruct}) + HippoRAG2}} \\ \midrule
Entity-KG (Llama-3-8B-Instruct)  & 48.33 & 72.58 & 67.34 & 84.25 & 77.59 & 92.16 \\
Entity-Event-KG (Llama-3-8B-Instruct) & 48.83 & 72.7 & 68.59 & 85.85 & 81.26 & 92.66 \\
Full-KG (Llama-3-8B-Instruct)  & {{49.12}} & {72.48} & {68.46} & {84.6} & {84.17} & {93.04} \\ \bottomrule
\end{tabular}

\caption{Recall @ 2 and Recall @ 5. }
\label{tab:recall_performance}
\caption{Recall performance in the knowledge graph created by Llama-3-8B-Instruct shows strong performance that is comparable with the knowledge graph created with 70B model.}
\end{table*}

\section{Details and Full Results on General Benchmarks}

\subsection{Implementation and Evaluation Details on FELM}
\label{sec:appendix_felm}
For the evaluation metrics, we follow the original paper~\cite{chen2023felm} and use balanced accuracy and F1 score to evaluate the factuality checking capability. For the classification of segments in an instance, we ask the model to generate the ID of false segments, and then get the true positive (TP), false positive (FP), true negative (TN) and false negative (FN) results. The balanced accuracy is calculated as:
\begin{equation}
    \small
    \text{Balanced Accuracy} = \frac{TP}{TP + FN} + \frac{TN}{TN + FP}
\end{equation}
Since we use F1 score to evaluate the factual error detection capability, we calculate the F1 score as:
% 2*(TN/(TN+FP))*(TN/(TN+FN))/(TN/(TN+FP)+TN/(TN+FN))
\begin{equation}
    \text{F1} = \frac{2 \cdot \text{Precision} \cdot \text{Recall}}{\text{Precision} + \text{Recall}}
\end{equation}
where Precision = $\frac{TN}{TN+FN}$ and Recall = $\frac{TN}{TN+FP}$.

We use the Retrieval-Augmented Generation method with different knowledge bases on 3 domains (world knowledge, science and technology, and writing/recommendation) of FELM benchmark. For the math domain and reasoning domain, we use the vanilla setting and their results are the same across different knowledge bases. The detailed results of the 3 domains are shown in Table~\ref{tab:full_fact_results}.
\begin{table*}[t]
\centering
\small
\begin{adjustbox}{max width=1\linewidth}
{
    \begin{tabular}{l|l|cccc|cccc|cccc}
        \toprule
        \multirow{2}{*}{\textbf{Corpus}} & \multirow{2}{*}{\textbf{Method}} & \multicolumn{4}{c|}{\textbf{World Knowledge}} & \multicolumn{4}{c|}{\textbf{Science and Technology}} & \multicolumn{4}{c}{\textbf{Writing/Recommendation}} \\
        \cmidrule{3-14}
        & & P & R & F1 & Acc & P & R & F1 & Acc & P & R & F1 & Acc \\
        \midrule
        - & - & 36.67 & 29.93 & 32.96 & 55.10 & 15.43 & 24.51 & 18.94 & 50.49 & 25.95 & 12.73 & 17.09 & 52.69 \\
        \midrule
        \rowcolor[gray]{0.9} \multicolumn{14}{l}{\textit{Text Corpora}} \\
        \midrule
        \multirow{3}{*}{Wikipedia} & Random &  31.78 & 27.89 & 29.71 & 52.52 & 7.95 & 11.76 & 9.49 & 43.94 & 21.95 & 26.97 & 24.20 & 53.78\\
        & BM25 & 26.82 & 32.65 & 29.45 & 49.31 & 15.23 & 38.24 & 21.79 & 50.48 & 29.21 & 48.69 & 36.52 & 62.40  \\
        & Dense Retrieval &  33.93 & 38.78 & 36.19 & 54.97 & 16.92 & 43.14 & 24.31 & 53.01 & 25.22 & 43.45 & 31.91 & 58.68  \\
        \midrule
         \multirow{3}{*}{Pes2o-Abstract} & Random &  27.36 & 19.73 & 22.92 & 49.86 & 10.13 & 15.69 & 12.31 & 45.64 & 26.92 & 28.84 & 27.85 & 56.50 \\
        & BM25 &  32.43 & 32.65 & 32.54 & 53.34 & 11.18 & 17.65 & 13.69 & 46.54 & 26.75 & 32.96 & 29.53 & 57.34\\
        & Dense Retrieval &  31.43 & 29.93 & 30.66 & 52.50 & 20.51 & 47.06 & 28.57 & 57.55 & 25.37 & 32.21 & 28.38 & 56.51 \\
        \midrule
        \multirow{3}{*}{Common Crawl} & Random &  30.14 & 29.93 & 30.03 & 51.72 & 11.17 & 21.57 & 14.72 & 45.75 & 23.73 & 28.09 & 25.73 & 54.91\\
        & BM25 & 27.21 & 27.21 & 27.21 & 49.71 & 12.21 & 25.49 & 16.51 & 46.68 & 27.32 & 40.82 & 32.73 & 59.42\\
        & Dense Retrieval & 25.50 & 34.69 & 29.39 & 48.00 & 15.48 & 36.27 & 21.70 & 50.78 & 24.41 & 38.95 & 30.01 & 57.27 \\
        \midrule
        \rowcolor[gray]{0.9} \multicolumn{14}{l}{\textit{Knowledge Graph}} \\
        \midrule
        Freebase & Think on Graph & 26.00 & 8.84 & 13.20 & 49.62 & 23.76 & 23.53 & 23.65 & 55.15 & 42.86 & 12.36 & 19.19 & 54.51 \\
        \midrule
        ATLAS-Wiki &  & 33.33 & 42.18 & 37.24 & 54.98 & 16.45 & 50.00 & 24.76 & 52.75 & 32.82 & 32.21 & 32.51 & 59.43  \\
        ATLAS-Pes2o & HippoRAG2 &  39.17 & 31.97 & 35.21 & 56.51 & 21.35 & 40.20 & 27.89 & 57.13 & 30.37 & 15.36 & 20.40 & 54.11 \\
        ATLAT-CC &  & 33.80 & 48.98 & 40.00 & 56.18 & 18.06 & 52.94 & 26.93 & 55.42 & 24.20 & 25.47 & 24.82 & 54.66  \\
        \bottomrule
    \end{tabular}
}
\end{adjustbox}
\caption{Factuality results (\%) on different domains of FELM benchmark with different Text Corporas and retrieval methods. P, R, F1, and Acc denote Precision, Recall, F1 score, and Balanced Accuracy, respectively.}
\label{tab:full_fact_results}
\end{table*}

\subsection{Implementation and Evaluation Details on MMLU}

\label{sec:appendix_mmlu}
Table \ref{tab:subject_classification} presents our classification for organizing MMLU tasks into distinct subject categories, providing a structured framework for domain-specific performance analysis. Table \ref{tab:general_benchmarks_ful} displays comprehensive results across all MMLU subject areas, revealing an important insight: while retrieval-augmented generation enhances performance in knowledge-intensive domains, it can negatively impact performance on reasoning-focused tasks such as mathematics and logical reasoning. This finding aligns with previous research suggesting that RAG may sometimes interfere with LLMs' inherent reasoning capabilities.

\begin{table*}[h]
        \small
	\centering
	\begin{tabularx}{\textwidth}{l|X}
		\toprule
        Subject&Task \\
        \midrule
        History&high school european history, high school us history, high school world history, prehistory \\
        \cmidrule{1-2}
	Formal Logic & formal logic, logical fallacies \\
        \cmidrule{1-2}
        Law& international law, jurisprudence, professional law \\
        \cmidrule{1-2}
        Philosophy and Ethics& philosophy, moral disputes, moral scenarios, business ethics\\
        \cmidrule{1-2}
        Religion&world religions\\
        \cmidrule{1-2}
        Medicine and Health& clinical knowledge, college medicine, medical genetics, professional medicine, virology, human aging, nutrition, anatomy\\
        \cmidrule{1-2}
        Social Sciences& high school geography, high school government and politics, high school psychology, professional psychology, sociology, human sexuality, us foreign policy, security studies\\
        \cmidrule{1-2}
        Economics& high school macroeconomics, high school microeconomics, econometrics\\
        \cmidrule{1-2}
        Business and Management& management, marketing, professional accounting, public relations\\
        \cmidrule{1-2}
        Math& abstract algebra, college mathematics, elementary mathematics, high school mathematics, high school statistics\\
        \cmidrule{1-2}
        Natural Sciences& astronomy, college biology, college chemistry, college physics, conceptual physics, high school biology, high school chemistry, high school physics\\
        \cmidrule{1-2}
        Computer Science and Engineering& college computer science, high school computer science, computer security, electrical engineering, machine learning\\
        \cmidrule{1-2}
        Global Facts& global facts, miscellaneous\\
        \bottomrule
	\end{tabularx}
     \vspace{-0.1in}
	\caption{The correspondence between subjects and tasks.}
	\label{tab:subject_classification}
\end{table*}

\begin{table*}[]
\huge
\centering
\resizebox{\linewidth}{!}{
\begin{tabular}{@{}l|c|ccccccccccccc@{}}
\toprule
\multirow{2}{*}{Corpus} 
% & \multirow{2}{*}{Method} 
& \multicolumn{13}{c}{MMLU} 
% & \multirow{2}{*}{MMLU\_pro} & \multirow{2}{*}{BBH} 
\\ \cmidrule(l){2-15}
   & \multicolumn{1}{c|}{overall}& \multicolumn{1}{c}{History} & \multicolumn{1}{c}{Law} & \multicolumn{1}{c}{Religion} & \multicolumn{1}{c}{PaE} & \multicolumn{1}{c}{MaH} & \multicolumn{1}{c}{GF} & \multicolumn{1}{c}{BaM} & \multicolumn{1}{c}{SS} & \multicolumn{1}{c}{Logic} & \multicolumn{1}{c}{Econ} & \multicolumn{1}{c}{Math} & \multicolumn{1}{c}{NS} & \multicolumn{1}{c}{CSaE}\\ \midrule
 None &69.18&76.59 &66.86&83.04& 63.55& 70.38& 66.72 & 72.20& 79.74& 64.35&\underline{68.35}&57.31&65.27&66.70\\
 Freebase-ToG &\textbf{70.36}&\textbf{78.42} &69.00&75.44& \underline{65.67}& \underline{72.65}& 67.27 & \textbf{73.67}& 76.00& 66.03&67.34&60.56&69.39&\textbf{68.23}\\
\midrule
\rowcolor[gray]{0.9} \multicolumn{15}{l}{\textit{Random Baseline}} \\
\midrule
Wikipedia & 68.06& 76.64& 66.82&79.53& 59.26& 70.34& 66.46& 67.34& 77.78&59.21 & 65.35& 60.91&67.52 &61.87\\
 Common Crawl & 67.93&74.89 & 66.52&79.53& 61.74& 69.82& 68.11 & 67.67& 77.52&59.30& 64.20&59.80& 67.42&62.22\\
 Pes2o-Abstract &68.07& 76.24& 64.16&80.70 &62.01 & 70.62& 66.59 & 69.27& 77.16&62.39& 64.70& 60.07& 66.41&62.18\\
\midrule
\rowcolor[gray]{0.9} \multicolumn{15}{l}{\textit{Text Corpora + DBM25}} \\
\midrule
 Wikipedia &68.99&76.67 & 67.35&78.36&63.34 &69.35 & 61.98& 71.39&76.99& 62.30&65.56 & 61.67 & 69.31&65.60\\
 Common Crawl&68.33 &76.15 & 66.36& 80.12& 60.43 & 69.58& 64.67 &69.47 &76.71&63.18& 68.22& \underline{62.26}& 65.55&65.04\\
Pes2o-Abstract & 69.04&78.01& 65.89 & 78.95&63.83 &71.01 & 65.78 & 68.29&77.07 &59.34 &\textbf{68.87} & 61.20& 67.73&65.74\\
\midrule
\rowcolor[gray]{0.9} \multicolumn{15}{l}{\textit{Text Corpora + Dense Retrieval}} \\
\midrule
Wikipedia &\underline{69.37}&73.59 & \underline{69.60}&79.53 & 63.58& 70.82& 62.41 & \underline{72.57}&76.83& 62.21& 67.35& 61.79&\textbf{69.81} &65.39\\
Common Crawl&67.03 &74.47 &68.98 &79.53 &60.46 &69.29 & 64.09 & 68.88&75.21&61.86& 62.27& 57.13& 64.54&64.47\\ 
Pes2o-Abstract & 69.07&75.79 &61.82 & 78.36& 65.15& 69.72& 66.77& 69.02& 76.47 & 63.05 & 63.07& \textbf{63.92}& \underline{69.53}&\underline{67.86}\\  
\midrule
\rowcolor[gray]{0.9} \multicolumn{15}{l}{\textit{ATLAS + HippoRAG2 }} \\
\midrule
ATLAS-Wiki &68.22 & 76.73& 67.38& \underline{84.21}& \textbf{66.01}& 70.82& \underline{68.36} & 72.35&79.16&63.65 & 64.53& 50.09& 65.45&62.10\\ 
ATLAS-CC&68.26 & \underline{78.16}& \textbf{70.85}& 83.04&65.60& 71.28&  63.95& 68.84& 78.16&65.42& 67.51& 52.87& 63.32&63.40\\
ATLAS-Pes2o&69.19 & 77.13&68.41& 81.29& 65.05&\textbf{72.75}  & 65.67 &72.32 & \underline{81.19}& 62.98&65.29& 54.24& 65.41&64.04\\ 
\midrule
\rowcolor[gray]{0.9} \multicolumn{15}{l}{\textit{ATLAS + ToG}} \\
\midrule
ATLAS-Wiki  &68.29& 77.91&66.60&\underline{84.21}& 65.10& 70.69& 63.85&70.49&78.31&\underline{67.08}& 66.24&54.41&63.65& 64.18\\ 
ATLAS-CC& 68.40&77.07&68.18&83.63&65.24& 72.03& 66.87&71.21&79.72& 66.59 &66.42&48.74&63.97&64.22\\
ATLAS-Pes2o&68.97&77.52&66.95&\textbf{84.80}& 63.44&71.15&\textbf{68.92}&69.98&\textbf{81.59}&\textbf{67.87}& 66.17& 55.46&63.35& 64.75\\ 
\bottomrule
\end{tabular}}
\caption{Performance comparison of our knowledge graph (KG) integrated with HippoRAG2 and ToG against baseline retrieval methods (Random, BM25, Dense Retrieval) across Wikipedia, Common Crawl, and Pes2o-Abstract corpora on MMLU benchmarks. Tasks are classified according to subjects, with bold and underline indicating the highest and the second highest performance. PaE, MaH, GF, BaM, SS, Econ, NS and CSaE represent Philosophy\_and\_Ethics, Medicine\_and\_Health, Global\_Facts, Business\_and\_Management, Social\_Sciences, Economics, Natural\_Sciences and Computer\_Science\_and\_Engineering respectively.}
\label{tab:general_benchmarks_ful}
\end{table*}

% \section*{Acknowledgements}
%  We thank Huawei for the support and computing resources from HKUST. 

\end{document}